\documentclass{article}
\PassOptionsToPackage{table,dvipsnames}{xcolor}
\usepackage{iclr2027_conference,times}

\usepackage[T1]{fontenc}
\usepackage{microtype}
\usepackage{inconsolata}
\usepackage{graphicx}
\usepackage{booktabs}
\usepackage{multirow}
\usepackage{amsmath}
\usepackage{amssymb}
\usepackage[ruled]{algorithm}
\usepackage{algpseudocode}
\usepackage[most]{tcolorbox}
\tcbuselibrary{listings}
\usepackage{placeins}
\usepackage{hyperref}
\usepackage{url}

\graphicspath{{figs/}}

\title{FinCutBench: A Continually Updated Benchmark for Measuring the Temporal Boundaries of LLM Financial Knowledge}

\author{
Siyuan Li\textsuperscript{1}
\quad Ke Wan\textsuperscript{1}
\quad
Boyong Hou\textsuperscript{1}
\quad Yujie Deng\textsuperscript{1}
\quad Zixuan Yuan\textsuperscript{1,\textdagger}
\\[6pt]
{\small \textsuperscript{1}The Hong Kong University of Science and Technology (Guangzhou)}
\\[3pt]
{\small \textsuperscript{\textdagger}Corresponding author}
\\[4pt]
{\small \texttt{sli974@connect.hkust-gz.edu.cn}}
\quad
{\small \texttt{zixuanyuan@hkust-gz.edu.cn}}
}

\iclrpreprintcopy
\begin{document}
\raggedbottom
\maketitle

\begin{abstract}
In financial backtesting and other point-in-time tasks, external inputs can be restricted by date, but model parameters may retain financial facts disclosed later, introducing parametric look-ahead bias. We introduce \textbf{FinCutBench}\footnote{Project repository: \url{https://github.com/FinCutBench/FinCutBench}.}, a continually updated monthly financial knowledge benchmark for measuring how far a fixed model's financial knowledge extends over calendar time. The current dataset contains 9,000 Chinese and English four-option multiple-choice questions from January 2023 to September 2026, covering Chinese and U.S. markets and four categories of financial information. We construct monthly knowledge profiles using the approximate months in which the tested facts became public. After a profile passes the decline test and knot checks, we define the FinCut Boundary as the first month when the fitted profile reaches its lower level, and report the stability of the estimated date under resampling. Boundary dates and localization precision vary across models. Using the independently measured Transition Start and Boundary as fixed references for downstream sample grouping, we find significant relative declines in U.S. asset-ranking performance and in earnings recall for Sol and Luna. Downstream effects vary across markets and tasks. The FinCut Boundary thus provides an empirical reference for the temporal scope of parametric knowledge in point-in-time financial evaluation, supporting evaluation-window selection and result interpretation.
\end{abstract}

\section{Introduction}

Large language models have been used for financial question answering, prediction, the construction of text-based variables, and trading agent evaluation~\cite{finqa,tatqa,financebench,finben,investorbench,fintradebench}. Model parameters may retain market movements, earnings results, and policy information disclosed after the time being evaluated. This issue is particularly important in financial backtesting and other point-in-time tasks: researchers can restrict news, market data, and financial data by date, but models may still draw on later knowledge stored in their parameters, creating parametric look-ahead bias. Prior studies show that this knowledge can change stock prediction signals, backtest returns, and financial text indicators~\cite{memorizationforecast,datedgpt,profitmirage,fincad2026,goalmissingtruth}.

Input materials have dates that can be checked, but the temporal scope of knowledge stored in model parameters is usually not directly visible. Vendor disclosure dates vary in scope and definition, making them difficult to use for consistent comparisons of financial knowledge coverage across specific model versions~\cite{dateddata,llmlagbench}. Financial knowledge coverage also varies by year, company characteristics, and market attention~\cite{historicalfinanceqa}. There is therefore a lack of reusable, continually updated resources for measuring the temporal scope of financial knowledge. We study which calendar months a fixed model's financial knowledge extends to, and whether this temporal scope can provide an empirical time reference for point-in-time financial evaluation.

We introduce FinCutBench, which combines a continually updated bank of financial fact questions with measurement of the temporal scope of a fixed model's knowledge. The question bank is organized by month according to when information becomes public and the rules for each source type. It covers the Chinese and U.S. markets in Chinese and English, respectively, with four information axes: macroeconomics, policy, and public affairs; sectors and supply chains; market paths; and individual asset moves. Questions are drawn from financial time series and securities research reports and use a common format of fact-based multiple-choice questions with four options. The current question bank contains 9,000 questions over 45 months, from January 2023 to September 2026. Continually adding new facts extends the measurement to more recent months as models evolve. This follows the approach of dynamic evaluation, in which questions are updated as facts change~\cite{livebench,realtimeqa,freshqa,cdqa}.

For a given model version, we aggregate closed-book performance by the month in which each fact first became public, producing a monthly knowledge profile over calendar time. FinCut Boundary is the first month at which this profile reaches the later low plateau after passing the overall decline test and knot checks. The piecewise curve also gives Transition Start, the month when the high plateau ends and the decline begins. Our main results report decline detection, the Boundary point estimate, and resampling-based localization stability together. Results by information axis further show temporal differences across types of financial information.

\begin{figure*}[!tp]
    \centering
    \includegraphics[width=\textwidth]{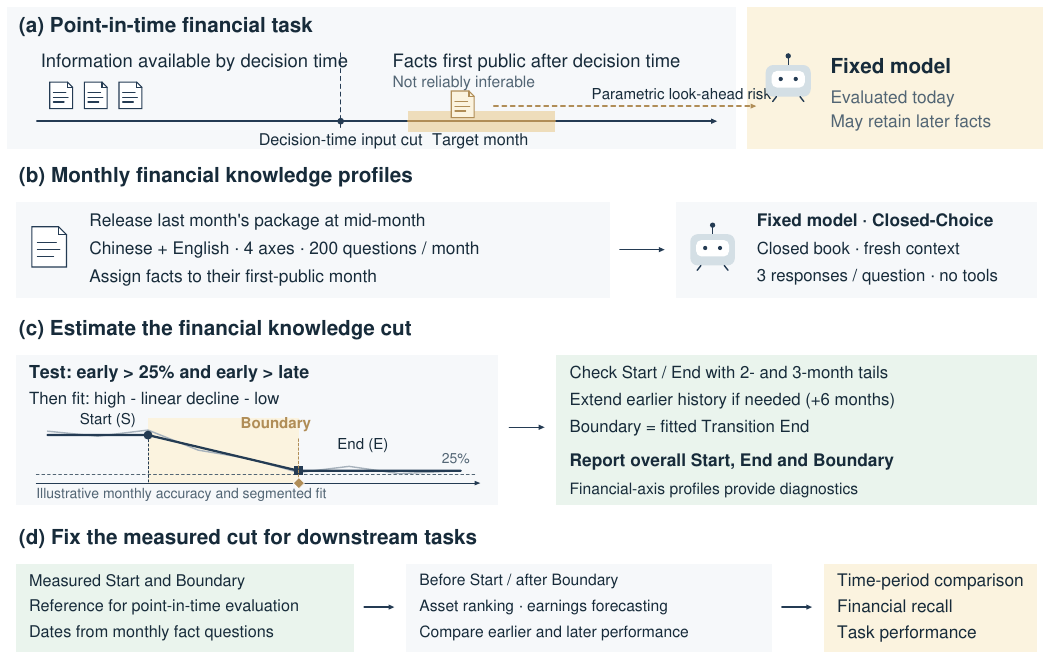}
    \caption{\textbf{Overview of FinCutBench.} Point-in-time tasks restrict external inputs to the decision time, while a fixed model may retain facts first disclosed later. Monthly financial questions characterize this parametric knowledge over calendar time. We test for an overall accuracy decline, fit its Transition Start and FinCut Boundary, report localization stability, and evaluate the measured dates as time references in downstream tasks.}
    \label{fig:fincutbench_overview}
\end{figure*}

We use the overall Start and Boundary independently measured by factual question answering as fixed time references for before-and-after comparisons in asset return ranking and earnings recall. The quarterly earnings task also measures earnings per share (EPS) prediction from materials available at the time, reusing the same company-quarters and result-disclosure periods as the recall analysis to test whether changes in recall are accompanied by changes in forecasting. Figure~\ref{fig:fincutbench_overview} shows the overall workflow connecting point-in-time input constraints, monthly knowledge measurement, and downstream comparisons.

Our contributions consist of three connected components: a financial fact resource updated monthly across two markets and four information axes; a measurement protocol that characterizes the temporal scope of a fixed model's financial knowledge through monthly knowledge profiles, a unified Boundary definition, and localization stability; and an evaluation of the Boundary's role as a time reference in asset ranking and quarterly earnings tasks.

\section{Related Work}
\label{sec:related_work}

\subsection{Parametric Look-Ahead Bias in Financial Backtesting}

Look-ahead bias in financial backtesting and historical point-in-time prediction can arise from model parameters that memorize outcomes disclosed later. \citet{detectinglookahead} construct outcome-recall queries for specific companies and dates and analyze the relationship between this memory and predictions of stock returns and capital expenditure. Profit Mirage combines question answering about historical financial facts, trading evaluation, and counterfactual experiments to diagnose information leakage~\cite{profitmirage}. Related control and mitigation methods target different stages: DatedGPT restricts the time range of training data, FinCAD uses adaptive decoding at inference time to reduce parametric look-ahead bias, and DeepFund uses live market evaluation~\cite{datedgpt,fincad2026,deepfund2025}.

\subsection{Temporal Knowledge Coverage}

Dated Data estimates effective cutoffs at the resource level using historical resource versions, while LLMLagBench identifies temporal knowledge boundaries from changes in scores on dated questions~\cite{dateddata,llmlagbench}. In finance, Beyond the Reported Cutoff examines how knowledge of company financials varies by year and company characteristics~\cite{historicalfinanceqa}. HindsightBench estimates effective cutoffs using date and outcome-recall probes within a task and applies its audit protocol to cross-domain checks~\cite{hindsightbench}.

FinCutBench uses a monthly bank of financial fact questions as a shared measurement resource across models. It forms monthly knowledge profiles, estimates Transition Start and FinCut Boundary after detecting an overall accuracy decline, and separately evaluates date stability. Appendix~\ref{sec:appendix_extended_related_work} provides broader context on temporal knowledge measurements and their downstream uses, including HindsightBench.

\section{Problem Setup}
\label{sec:task_definition}

FinCutBench measures a fixed model version's closed-book performance on financial facts from different months to characterize the temporal scope of its knowledge. The model receives only the question, answer options, and response instructions, and answers using its own knowledge. This measurement provides point-in-time financial tasks with a time reference for the knowledge stored in model parameters.

Each monthly question set covers one complete calendar month. Questions are assigned to months based on when the facts became public and the rules for each source type. They target facts newly available in the target month that are difficult to infer reliably from earlier information. For example, the actual interest rate decision announced at a monetary policy meeting is a new fact for that month and is suitable for assessing the model's knowledge of that month's information. We summarize accuracy and statistical uncertainty across consecutive months as the model's \textbf{monthly knowledge profile}. This profile describes the model's performance on financial facts from each month under the current question bank and inference configuration. Source processing and question inclusion criteria are described in Section~\ref{sec:benchmark_construction}.

\subsection{FinCut Boundary}
\label{sec:boundary_definition}

\textbf{FinCut Boundary is the first month at which the monthly knowledge profile reaches the later low plateau after passing the overall decline test and knot checks.}

After detecting a significant decline in overall accuracy, we fit the monthly knowledge profile with three phases: a high plateau, a linear decline, and a low plateau. \textbf{Transition Start} is the month marking the end of the high plateau and the start of the decline. Both plateau levels are estimated from the data, so the Boundary summarizes the temporal scope of the model's knowledge of the tested financial facts based on its observed behavior.

At Start, the fitted curve is still at the high level; at Boundary, it has reached the low level. When the two dates fall in adjacent months, the decline occurs between those months, with no full intermediate month.

The main results estimate the overall FinCut Boundary from all questions and use it as a common time reference for model knowledge in downstream experiments. For tasks that focus on a specific type of information, the Boundary of the corresponding information axis provides additional diagnostics. For example, macroeconomic tasks concern the macro/policy axis, and sector analysis concerns the sector/supply-chain axis. For each of the four information axes, we report monthly profiles, Boundaries that independently pass the checks, and unresolved cases. The decline test and knot checks determine the overall Boundary point estimate; Localization Rate and Boundary Range describe how sensitive this date is to the question sample. A Leading-Axis Boundary requires the corresponding axis to have a Boundary Range of at most six months. Details are given in Section~\ref{sec:boundary_framework}.

\section{Benchmark Construction}
\label{sec:benchmark_construction}

\subsection{Dataset Coverage}

FinCutBench organizes financial facts first made public in each calendar month into a separate monthly question set. It defines four axes around the information needed for financial analysis: macroeconomics, policy, and public affairs; sectors and supply chains; market paths; and individual asset moves. These axes cover economic and policy conditions, sector supply and demand, market-wide price dynamics, and asset-specific movements, respectively. Each month contains 200 fact-based multiple-choice questions with four options, with 50 questions per axis and equal numbers in Chinese and English. The data version used in this study contains 45 monthly question sets from January 2023 to September 2026, covering the Chinese and U.S. markets in Chinese and English, respectively, for a total of 9,000 questions. Model measurements use consecutive observation windows selected from these months.

Two types of sources complement each other. Indicator and market data support numerical and market-path questions, whose answers can be recomputed and verified from the original data and calculation rules. Securities research reports provide policy, public affairs, and sector events that require detailed context. Chinese time series come from Tonghuashun, U.S. structured indicators come from Federal Reserve Economic Data (FRED), and U.S. market data are obtained through yfinance. Research reports are purchased through third-party channels. Each month follows the source and information-axis proportions in Table~\ref{tab:benchmark_composition} to reduce the effect of composition changes on temporal comparisons. Indicators, assets, and representative questions are listed in Appendix~\ref{sec:appendix_benchmark_details}.

\begin{table}[!htbp]
\centering
\begin{minipage}[c]{0.64\textwidth}
\centering\footnotesize
\setlength{\columnwidth}{\linewidth}
\begingroup
\renewcommand{\arraystretch}{1.15}
\setlength{\tabcolsep}{3pt}
\begin{tabular*}{\columnwidth}{@{\extracolsep{\fill}}lrrrr@{}}
\toprule
Axis & Reports & Indicators & Market & Total \\
\midrule
Macro / policy & 30 & 20 & 0 & 50 \\
Sector / supply & 40 & 10 & 0 & 50 \\
Market path & 0 & 0 & 50 & 50 \\
Asset moves & 0 & 0 & 50 & 50 \\
\midrule
\textbf{Total} & \textbf{70} & \textbf{30} & \textbf{100} & \textbf{200} \\
\bottomrule
\end{tabular*}
\endgroup

\end{minipage}\hfill
\begin{minipage}[c]{0.32\textwidth}
\caption{Monthly question composition across both markets and languages.}
\label{tab:benchmark_composition}
\end{minipage}
\end{table}

\subsection{Question Construction}

Figure~\ref{fig:fincutbench_construction} in the appendix summarizes question generation, quality review, and monthly question set construction. Each question is assigned to a month based on when the tested fact became public and the timing rules for its source type.

\textbf{Questions from time-series data.} U.S. indicator questions use historical FRED data vintages available by the end of the question's assigned month. For Chinese monthly indicators, the question set for month $t$ uses only observations from $t-1$ or earlier, treating observations for $t-1$ as available by the end of month $t$. When historical vintages are unavailable, database snapshots are used for a fixed set of series with low revision risk. Market-data questions use data from complete trading intervals, and numerical answers are checked through independent calculations. Distractors are plausible numerical values with matching units and precision, or valid trading dates within the same window.

\textbf{Questions from research reports.} The generation agent reads the full relevant context in materials from the target month, first identifies the facts and the evidence for their timing, and then creates a self-contained question, a correct answer, three distractors, and an explanation. Distractors come from the same financial context and are mutually exclusive and incorrect under the conditions stated in the question. An independent quality-control agent uses a fresh context, rereads the evidence, and checks factual status, temporal conditions, answer uniqueness, and the plausibility of each distractor in the corresponding financial setting. Revised and replacement questions undergo the full checks again.

\subsection{Quality Validation}

All questions undergo five content checks. \textbf{Temporal validity} requires the assigned month to match when the tested fact became public, following the timing rules for each source type. \textbf{Financial relevance} requires the fact to be relevant to financial analysis and judgment. \textbf{Dependence on new information} requires correct answers to depend mainly on information newly available in the target month. \textbf{Evidence sufficiency} requires identifiable source materials that directly support the answer and its necessary conditions. \textbf{Question wording and option design} require each question to include all necessary information and have a unique correct answer. The four options must be consistent in type, units, and wording, and the question and options must not provide obvious guessing cues. The full review criteria are provided in Appendix~\ref{sec:appendix_question_criteria}.

Monthly question sets are deduplicated by fact and maintain the same proportions across sources and information axes to reduce the effect of composition changes on temporal comparisons. Section~\ref{sec:experiments} further analyzes difficulty changes across months using answers from different models to the same questions.

\newcommand{\HumanInstitutionExpertN}{2}
\newcommand{\HumanFinanceDegreeExpertN}{2}
\newcommand{\HumanFinalN}{126}
\newcommand{\HumanPassN}{126}
\newcommand{\HumanPassPct}{100.00\%}
\newcommand{\HumanWilsonLo}{97.04\%}
\newcommand{\HumanWilsonHi}{100.00\%}
\newcommand{\HumanUnanimousN}{122}
\newcommand{\HumanUnanimousPct}{96.83\%}
\newcommand{\HumanPairPct}{97.88\%}
\newcommand{\HumanKappa}{-0.011}
\newcommand{\HumanInitialN}{42}
\newcommand{\HumanInitialIssueN}{0}
\newcommand{\HumanFlagN}{1}
\newcommand{\HumanPrecisionPct}{0.00\%}
\newcommand{\HumanChangedN}{3}
\newcommand{\HumanChangedPct}{7.14\%}
\newcommand{\HumanStemChangedN}{1}
\newcommand{\HumanOptionsChangedN}{2}
\newcommand{\HumanUnchangedN}{39}
\newcommand{\HumanUniqueN}{129}
\newcommand{\HumanDisagreementN}{4}
\newcommand{\HumanRegressionN}{0}
\newcommand{\HumanMajorityFailureN}{0}
\newcommand{\HumanExpertMinPct}{98.41\%}
\newcommand{\HumanExpertMaxPct}{99.21\%}

Human review primarily evaluates whether the final questions are ready for direct use in benchmark evaluation and also checks content changes before and after independent quality control. The three representative months are January 2023, March 2025, and September 2026. Combined with the two markets, they form six question sets. We draw stratified samples of report questions, which three experts independently review against the five criteria. A judgment shared by at least two experts determines whether each question meets the requirements. Among the three experts, \HumanInstitutionExpertN{} have worked at major financial institutions, and \HumanFinanceDegreeExpertN{} hold a bachelor's or master's degree in a finance-related field. Questions generated by different models are analyzed separately. The methods and full results are given in Appendices~\ref{sec:appendix_human_alignment_protocol} and~\ref{sec:appendix_human_alignment_results}, respectively.

All \HumanFinalN{} reviewed final questions generated by GPT-6.1-Sol-High received majority approval. The rate of questions judged ready for direct use was \HumanPassPct{} (Wilson 95\% confidence interval: \HumanWilsonLo{}--\HumanWilsonHi{}). Of these, \HumanUnanimousN{} received unanimous approval from all three experts, accounting for \HumanUnanimousPct{}. Complete criterion-level statistics and the small number of disagreements are reported in the appendix.

\subsection{Monthly Updates}
\label{sec:monthly_releases}

The project repository is the canonical location for question-bank updates and release information. Under the monthly update schedule, each question set covers the preceding month and is prepared in the middle or later part of the following month. For example, an October question set can include policies and events disclosed in October, market data from trading in October, and September economic data first released in October.

\section{Evaluation Protocol}
\label{sec:evaluation_protocol}

We estimate FinCut Boundary from knowledge profiles over consecutive months through decline detection, date estimation, and stability assessment. The decline test provides evidence of differences between periods, curve fitting and knot checks determine the Boundary point estimate, and resampling measures the date's sensitivity to the question sample. When no decline is detected or the knot checks fail, we extend the window toward earlier months. The overall Boundary point estimate and resampling stability are reported together.

\subsection{Closed-Book Measurement}

The main evaluation uses Closed-Choice, requiring the model to select one of four answer options. Each question uses a separate context containing only the question, options, and response instructions, without evidence, the reference answer, or history from other questions. This setup makes monthly performance mainly reflect the model's own knowledge accessible under the current inference configuration. Each question is answered independently three times. We first average correctness for each question, then aggregate question scores with equal weight. Each month pools 200 questions across Chinese and English, with 50 per axis. Statistical inference uses the aggregated question-level scores.

The initial window includes the tested model version's release month and the preceding 12 months, restricted to the months covered by our fixed version of the question bank. If the release-month question set is not yet complete, the window temporarily ends at the latest month in the question bank, and this is noted in the results. If any expected months, questions, or response runs are incomplete, the result remains Pending. We report boundary dates, monthly accuracy, and 95\% confidence intervals for each month to show the changes associated with the reported dates. IDK-Enabled, which allows abstention, is used for supplementary analysis in Appendix~\ref{sec:idk_results}.

\subsection{Boundary Estimation}
\label{sec:boundary_framework}

\textbf{Decline test.} Overall accuracy alone determines the main Boundary. Within each observation window, we enumerate split points between earlier and later periods, requiring earlier accuracy to exceed both the 25\% chance reference level and later accuracy. We use one-sided Hoeffding tests based on fact groups and apply Holm correction for multiple testing across all split points for the overall profile. The significance level allocated to window check $j$ is $\alpha_j=0.05/[j(j+1)]$, with 0.025 for the initial window and a total of 0.05 across all checks. Questions testing the same fact form a fact group. Under the assumption that dependence is allowed within groups and groups are independent, the probability of falsely detecting an overall decline for a single model is bounded by 5\% across the window search. Each of the four information axes has its own family of tests and independently undergoes quality checks within the overall window. The overall result depends only on overall accuracy.

\textbf{Knot localization.} After the decline test passes, we describe accuracy with three phases: an early high plateau, a linear decline, and a late low plateau. The high and low levels are both estimated from the data, and months have equal weight in the fit. The high and low phases must each contain at least two months. Among valid fits, we select the curve with the smallest squared error, obtaining both Start, the last month of the high plateau, and FinCut Boundary, the first month at the low level. We then change the minimum duration of the low phase to three months and check how each knot changes. This step reuses the same decline test. Formulas and tie-breaking rules are given in Appendix~\ref{sec:appendix_boundary_methods}.

\textbf{Boundary and stability.} FinCut Boundary is the first month at which the main fit reaches the later low plateau after passing the decline test and knot checks. The later low level is estimated from the data and may be above or below the 25\% chance reference level. We use 1,000 resampling replicates of the full procedure to report two measures of stability: \textbf{Localization Rate} is the proportion of replicates in which Boundary can be located again, and \textbf{Boundary Range} is the central 90\% range of Boundary dates among these replicates. A range spanning at most six months is labeled Tight; a wider range is labeled Wide. We report the point estimate alongside this precision label. Monthly 95\% intervals separately describe uncertainty in each month's accuracy. The full sampling design is provided in Appendix~\ref{sec:appendix_fbc}.

\subsection{Window Expansion}

For each complete window, we first test for an overall decline, then check whether Start and Boundary obtained under the two minimum low-phase durations each differ by at most one month. Start in both fits must lie beyond the left endpoint of its allowed range to ensure enough observations in the high phase. We retain a flag when Boundary reaches the latest allowed date. Once an overall decline is detected and the knot checks pass, we determine the Boundary point estimate in the current window and stop expanding.

If no decline is detected or the knot checks fail, we add six earlier months within the declared observation range, keep the end month unchanged, and repeat testing and fitting. If no decline is detected after reaching the earliest month in the observation range, the result is ND. If a decline is detected but the knot checks still fail, the result is Unresolved. Missing questions, months, or response runs within the range produce Pending. All four information axes share the overall stopping window. Each axis separately corrects for its candidate split points and independently checks knot consistency under the two minimum low-phase durations and the left-edge condition. An axis qualifies as a Leading-Axis candidate when its Boundary Range spans at most six months. We select the latest date among candidate axes and retain ties. Axes do not expand their windows separately, and their results serve only as diagnostics. We separately check whether downstream materials and responses cover the required before-and-after windows. Algorithm~\ref{alg:boundary_window_expansion} presents the measurement procedure.

\subsection{Experimental Setup}

The main evaluation includes DeepSeek-V4.1-Flash-1017~\cite{fincut_deepseek_v41,fincut_deepseek_v41_report}, Qwen3.8-27B~\cite{fincut_qwen38}, Qwen3.5-4B, 122B-A10B, and 397B-A17B~\cite{fincut_qwen35}, GPT-5.6-Sol~\cite{fincut_gpt56_sol}, GPT-6-Astra~\cite{fincut_gpt6_astra}, Gemini-3.8-Flash~\cite{googledeepmind2026gemini38flashcard}, and Claude-Haiku-4.5 and Claude-Haiku-5.5~\cite{fincut_claude_haiku45_api,fincut_claude_haiku55_api}. Models from selected families provide supplementary diagnostics.

The default settings are temperature 0 and a maximum output of 16,384 tokens. For models that support explicit reasoning, we use the lowest setting that enables it. Qwen models are deployed through vLLM. DeepSeek is accessed through its official API, GPT through the Codex runtime, and Gemini and Claude through OpenRouter. Model-specific inference settings are given in Appendix~\ref{sec:appendix_experimental_setup}.

\section{Financial Knowledge Profiles}
\label{sec:experiments}

This section reports the models' monthly knowledge profiles, overall FinCut Boundaries, and localization stability. Monthly curves show how knowledge coverage changes over calendar time, while the four information axes show differences across categories. Resampling and comparisons on the same questions assess date stability and the effect of question difficulty.

\subsection{FinCut Boundaries}

Table~\ref{tab:main_results} reports release dates, vendor disclosures, decline detection, overall FinCut Boundaries, localization stability, and Leading-Axis Boundaries. Official dates come from documentation for the tested model versions~\cite{fincut_gpt56_sol_api,fincut_gpt6_astra_api,googledeepmind2026gemini38flashcard,fincut_claude_haiku45_api,fincut_claude_haiku55_api}; Appendix~\ref{sec:appendix_vendor_cutoffs} provides a systematic comparison of the three types of disclosure. Observation windows for each model are listed in Appendix Table~\ref{tab:main_results_coverage}.

\begin{table}[!t]
\centering\small
\resizebox{\textwidth}{!}{
\begingroup
\renewcommand{\arraystretch}{1.12}
\setlength{\tabcolsep}{2.5pt}
\begin{tabular}{@{}p{3.2cm}p{1.6cm}p{2.1cm}p{1.1cm}p{1.7cm}p{3.2cm}p{1.8cm}@{}}
\toprule
Model & Release Date & \shortstack[l]{Official\\Knowledge\\Cutoff} & \shortstack[l]{Decline\\Detected} & \shortstack[l]{Overall\\FinCut\\Boundary} & \shortstack[l]{Localization\\Stability} & \shortstack[l]{Leading-Axis\\Boundary} \\
\midrule
\multicolumn{7}{@{}l}{\textbf{Open-Weight Models}} \\
\multicolumn{7}{@{}l}{\textit{DeepSeek family}} \\
DeepSeek-V4.1-Flash-1017\textsuperscript{a}\textsuperscript{d} & 2026-10-17 & - & Yes & 2026-06 & \shortstack[l]{96.2\% located\\2026-06--2026-07 (Tight)} & 2026-08 \\
\multicolumn{7}{@{}l}{\textit{Qwen family}} \\
Qwen3.8-27B & 2026-08-14 & - & Yes & 2026-03 & \shortstack[l]{100.0\% located\\2026-02--2026-06 (Tight)} & 2026-03 \\
Qwen3.5-4B\textsuperscript{d} & 2026-03-02 & - & No & ND & -- & ND \\
Qwen3.5-122B-A10B\textsuperscript{d} & 2026-02-24 & - & Yes & 2025-12 & \shortstack[l]{99.6\% located\\2025-02--2026-01 (Wide)} & 2025-03 \\
Qwen3.5-397B-A17B\textsuperscript{d} & 2026-02-16 & - & Yes & 2025-04 & \shortstack[l]{100.0\% located\\2025-02--2025-07 (Tight)} & 2025-04 \\
\addlinespace[3pt]
\multicolumn{7}{@{}l}{\textbf{Closed-Weight Models}} \\
\multicolumn{7}{@{}l}{\textit{GPT family}} \\
GPT-6-Astra & 2026-09-03 & 2026-04-30 & Yes & 2026-06 & \shortstack[l]{100.0\% located\\2026-06--2026-06 (Tight)} & 2026-07 \\
GPT-5.6-Sol\textsuperscript{d} & 2026-07-09 & 2026-02-16 & Yes & 2026-06 & \shortstack[l]{90.2\% located\\2026-03--2026-06 (Tight)} & 2026-06 \\
\multicolumn{7}{@{}l}{\textit{Gemini family}} \\
Gemini-3.8-Flash & 2026-09-02 & 2026-03\textsuperscript{c} & Yes & 2026-03 & \shortstack[l]{99.9\% located\\2026-03--2026-06 (Tight)} & 2026-03 \\
\multicolumn{7}{@{}l}{\textit{Claude family}} \\
Claude-Haiku-5.5\textsuperscript{a}\textsuperscript{d} & 2026-10-07 & 2026-06 & Yes & 2026-06 & \shortstack[l]{99.6\% located\\2026-03--2026-07 (Tight)} & 2026-08 \\
Claude-Haiku-4.5\textsuperscript{d} & 2025-10-15 & 2025-02 & Yes & 2025-02 & \shortstack[l]{89.6\% located\\2024-07--2025-04 (Wide)} & \shortstack[l]{Not\\identified\textsuperscript{e}} \\
\bottomrule
\end{tabular}
\endgroup
}
\caption[Overall and leading-axis FinCut Boundaries]{Overall and leading-axis FinCut Boundaries for pooled Chinese and English questions. Stability gives the Localization Rate and conditional middle 90\% Boundary Range; Tight spans at most six months. Reported leading axes: Macro, except GPT-5.6-Sol (Macro/Asset). Ties require Tight ranges. \textsuperscript{e} Start reaches the fixed window's left limit.
 \textsuperscript{a} marks a provisional release window; \textsuperscript{c} indicates that Gemini reports a January 2025 limit for some domains; \textsuperscript{d} marks an expanded observation window.}
\label{tab:main_results}
\end{table}

The main table shows three types of monthly knowledge profile: profiles with a stably located Boundary, profiles with a detected decline but more widely spread date estimates, and profiles with no decline detected over the full observation range. The Boundaries are June 2026 for DeepSeek, Sol, Astra, and Claude-Haiku-5.5; March 2026 for Qwen3.8 and Gemini; and April 2025 for Qwen3.5-397B. All of these models have a Tight Boundary Range. Qwen3.5-122B and Claude-Haiku-4.5 have point estimates of December and February 2025, respectively, but their Boundary Ranges are Wide. No significant decline is detected for Qwen3.5-4B over its full observation range from January 2023 to March 2026, and it is marked No and ND. The point estimates for Haiku 5.5 and 4.5 both match the vendors' disclosed cutoff months for general and reliable knowledge. These two types of date describe financial question-answering behavior and vendor statements about knowledge coverage, respectively.

Claude-Haiku-4.5 passes the overall decline test within the 19-month point-estimation window from April 2024 to October 2025. Its adjusted $p=0.00544$ is below the threshold of $0.00833$ for this window check. Both low-phase length settings give a Start of January 2025 and a Boundary of February 2025. Resampling over the full procedure uses responses from October 2023 to October 2025 and gives a Localization Rate of 89.6\% and a Boundary Range from July 2024 to April 2025, resulting in a Wide label. The vendor discloses a February 2025 cutoff for general and reliable knowledge and a July 2025 cutoff for the full training data; the official-date column in the main table uses the former~\cite{fincut_claude_haiku45_api}.

Qwen3.5-122B shows an overall decline, with a full-data point estimate of December 2025 and a Localization Rate of 99.6\%. Its Boundary Range extends from February 2025 to January 2026. This indicates that a decline is usually detected across different question samples, while the specific month at which the decline reaches the later low level is more sensitive to the sample. Appendix~\ref{sec:appendix_qwen35_profiles} discusses the date difference between 122B and 397B.

\subsection{Accuracy Profiles}

\begin{figure}[H]
\centering
\includegraphics[width=\textwidth]{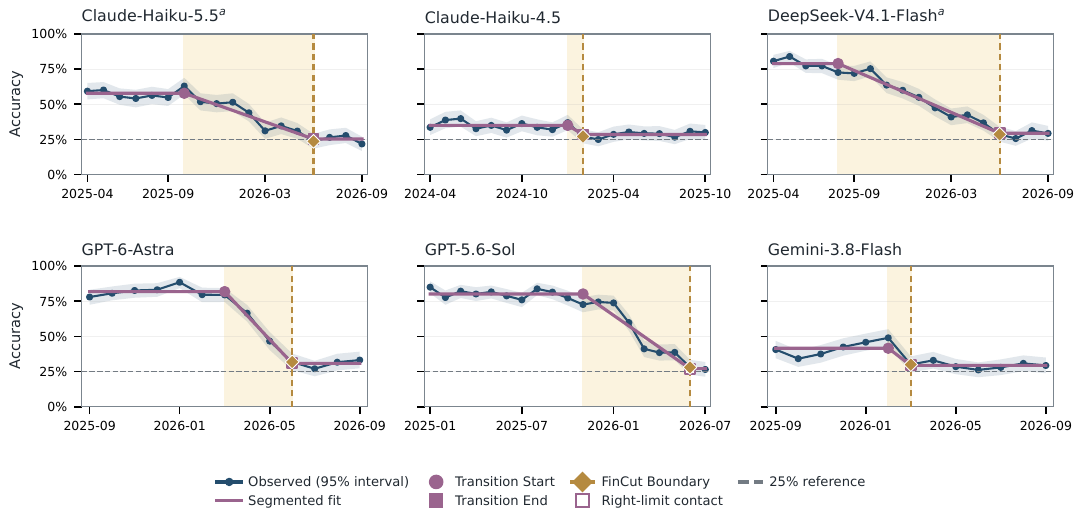}
\caption{\textbf{Monthly accuracy and overall FinCut Boundaries.} Each panel uses the model's stopping window. Blue shading shows pointwise 95\% fact-group bootstrap intervals; pale bands span [Start, Boundary]. \textsuperscript{a} marks provisional release windows.}
\label{fig:boundary_overall_profiles}
\end{figure}

Figure~\ref{fig:boundary_overall_profiles} shows monthly accuracy, pointwise intervals, fitted curves, and Boundaries for six models within their stopping windows. The top row contains Claude-Haiku-5.5, Claude-Haiku-4.5, and DeepSeek; the bottom row contains Astra, Sol, and Gemini. Haiku 4.5 drops from 37.0\% in January 2025 to 27.2\% in February and then fluctuates at a lower level. Haiku 5.5 begins to decline in October 2025 and has a FinCut Boundary of June 2026. The monthly trajectories and fitted knots together show how performance changes for each model.

\begin{figure}[H]
\centering
\includegraphics[width=\textwidth]{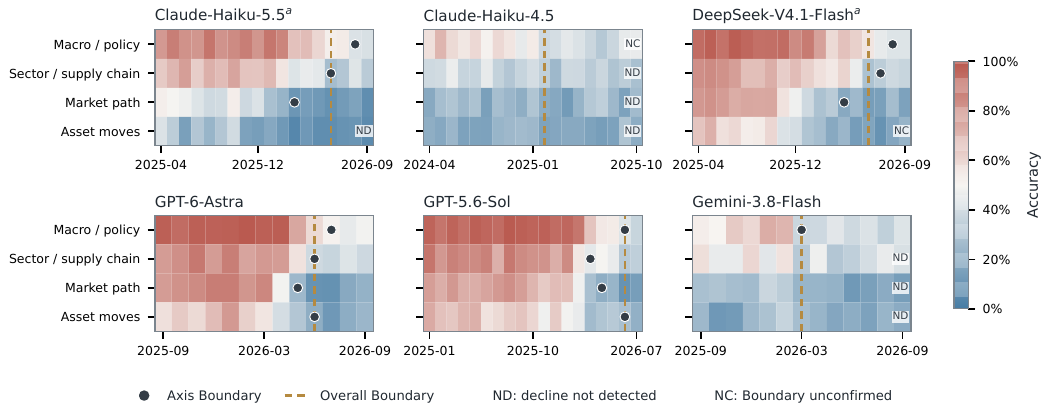}
\caption{\textbf{Financial-axis accuracy and axis-specific Boundaries.} Panels share the model order and observation windows of Figure~\ref{fig:boundary_overall_profiles}. \textsuperscript{a} marks provisional release windows.}
\label{fig:boundary_axis_heatmaps}
\end{figure}

Figure~\ref{fig:boundary_axis_heatmaps} uses the same model layout and observation windows to show accuracy on the four axes and the Boundary for each axis that passes the checks. Haiku 5.5 has Boundaries of March, June, and August 2026 on the market-path, sector/supply-chain, and macro/policy axes, respectively, with no decline detected on the asset-move axis. Haiku 4.5 has an overall Boundary of February 2025. Its macro/policy axis shows a decline, but the Start reaches the left limit; no decline is detected on the other three axes. Therefore, no dates are reported for its four axes.

\subsection{Validation Results}

All nine main models with a Boundary pass the knot checks within their selected windows. When the minimum duration of the low phase increases from two to three months, the Start and Boundary remain unchanged for eight models, while both knots move by one month for Sol.

We also run 1,000 resampling replicates of the full procedure for each model to assess whether the decline can be located again under different question samples and whether the resulting Boundary is close to the full-data point estimate. The nine main models with a Boundary have Localization Rates from 89.6\% to 100\%. Only 7.2\% of replicates yield a Boundary for Qwen3.5-4B, consistent with the absence of a detected decline in the full data. For Astra, 100\% of replicates yield a Boundary within one month of the point estimate, compared with 95.6\% for DeepSeek and 81.5\% for Haiku 5.5. For Haiku 4.5, 89.6\% of replicates yield a Boundary, 45.7\% exactly reproduce February 2025, and 74.3\% fall within one month of it. Qwen3.5-122B has a Localization Rate of 99.6\%, but only 38.2\% of replicates yield a Boundary within one month of December 2025, indicating that its date is sensitive to question sampling. The full distributions are reported in Appendix Table~\ref{tab:transition_pair_stability}.

Four earlier Qwen2.5 models of different sizes answer the same questions, with their average accuracy generally close to 25\%. For each of the five main models included in this comparison, the decline from the first four to the last four months of its default window is larger than that of the controls, and all pointwise 95\% confidence intervals for the difference are positive (Appendix~\ref{sec:appendix_construction_robustness}). When questions are separated by source, declines are also observed across questions produced by different construction methods. The next section examines whether the temporal scope of knowledge corresponds to period differences in downstream tasks.

\clearpage

\section{Downstream Time Diagnostics}
\label{sec:downstream}

This section examines whether dates obtained from factual measurements of model knowledge can serve as time references for window selection and result interpretation in point-in-time financial evaluation. Asset ranking and earnings recall directly compare performance before the Start and after the Boundary. Earnings forecasts reuse the disclosure-period groups from the recall analysis for paired comparisons.

\subsection{Shared Design}
\label{sec:downstream_design}

For each model with an overall FinCut Boundary, we fix its overall Start and Boundary to divide the downstream samples. Ranking compares up to six complete calendar months immediately before the Start with up to six complete calendar months immediately after the Boundary, excluding the knot months. These periods are called Pre-Start and Post-Boundary, respectively. Their date ranges are determined by question-bank coverage and the data cutoff. Models receive historical materials labeled with the observation date and task instructions, and answer each question in an independent context over three rounds.

The ranking experiments include Qwen3.8-27B, GPT-5.6-Sol, GPT-5.6-Luna, and Claude-Haiku-5.5. Qwen2.5-72B answers the same questions for the same months as each target model. The Chinese market uses Chinese and the U.S. market uses English, with results reported separately. Scores are first averaged across the three rounds within each month. We then compute the target model's difference from 72B in the same month, using months as the statistical units. Market data extend through September 2026. The GPT models and Haiku use six months before the Start and three months after the Boundary, while Qwen3.8 uses six months on each side. Overall knots and actual windows are listed in Appendix Table~\ref{tab:ranking_windows}.

The quarterly earnings experiments use the same four target models as ranking. Each model's Start and Boundary and the month in which the target quarter's results were first disclosed define Disclosure-Pre and Disclosure-Post. For earnings recall, these groups represent knowledge before and after the Boundary. The forecasting conditions reuse the same company-quarters, while their information cutoff remains the day before the target quarter begins. The analysis uses a fixed cohort of 50 companies for which all four models have eligible questions in both periods. The respective windows contain 190 company-quarter observations for Qwen3.8, 167 for Sol, and 162 each for Luna and Haiku. Qwen2.5-72B answers matching questions for the same companies, quarters, and windows as a reference for common changes in difficulty. Each condition uses three independent response rounds. Scores are first averaged within questions, then within each company and period, and finally across companies with equal weight.

\subsection{Asset Return Ranking}

Each market uses a fixed pool of 30 assets, comprising 15 stocks and 15 exchange-traded funds (ETFs). Each month, after the close of the last trading day of the preceding month, the model receives asset names, tickers, and cross-sectional percentiles of historical returns, volatility, drawdown, and changes in trading volume. It also receives recent actual returns and volatility for three representative indices in that market. Both types of information are presented as Markdown tables with values shown to two decimal places. Percentiles are computed separately within each market's 30-asset pool. The primary metric assumes that the ranking signal is implemented at the close of the first trading day of the current month, introducing a one-trading-day implementation lag. The target return is the adjusted return from that close to the close of the last trading day of the month. This setup draws on cross-sectional return prediction and portfolio evaluation~\cite{gu2020empirical}, with the input and output formats adapted to a zero-shot ranking task.

The model divides all 30 assets into high-, medium-, and low-expected-return groups of 10 assets each, with no scoring of the order within a group. It returns the groups using short integer IDs and provides a brief rationale. The use of IDs follows work on document reranking~\cite{sun-etal-2023-chatgpt,qin-etal-2024-large}. The primary metric is the Spearman rank correlation between the three-level group scores and realized returns, known as the Rank Information Coefficient (Rank IC). Higher values indicate more accurate rankings.

In Table~\ref{tab:downstream_main}, both Pre and Post are computed as the target model's Rank IC minus the same-month 72B Rank IC. A negative difference $\Delta=\mathrm{Post}-\mathrm{Pre}$ indicates a decline in relative performance. The primary tests examine the average change across the four equally weighted models separately for the Chinese and U.S. markets. Resampling is shared by calendar month to preserve correlations across models in the same month. One-sided Welch $t$ tests provide additional model-level results. All $p$ values are unadjusted; full methods are given in Appendix~\ref{sec:appendix_downstream}.

\begin{table*}[!tp]
\centering
\small
\begingroup%
\small%
\setlength{\tabcolsep}{2pt}%
\renewcommand{\arraystretch}{1.05}%
\begin{tabular*}{\textwidth}{@{\extracolsep{\fill}}lrrrrrr@{}}%
\toprule
Model & \multicolumn{3}{c}{China} & \multicolumn{3}{c}{US} \\
\cmidrule(lr){2-4}\cmidrule(lr){5-7}
& Pre & Post & Rel. $\Delta$ & Pre & Post & Rel. $\Delta$ \\
\midrule
Qwen3.8-27B & 0.106 & -0.038 & $-0.144^{*}$ & 0.054 & -0.261 & $-0.315^{***}$ \\
GPT-5.6-Sol & 0.147 & -0.031 & $-0.179$ & 0.129 & -0.326 & $-0.456^{**}$ \\
GPT-5.6-Luna & -0.011 & 0.063 & $+0.074$ & -0.108 & -0.328 & $-0.220$ \\
Claude-Haiku-5.5 & -0.027 & 0.017 & $+0.044$ & 0.114 & -0.219 & $-0.333^{**}$ \\
\bottomrule
\end{tabular*}%
\endgroup%

\caption{Asset ranking relative to same-month Qwen2.5-72B; Rel. $\Delta=\mathrm{Post}-\mathrm{Pre}$. Stars denote raw one-sided model-level $p$ values: $^{*}p<0.10$, $^{**}p<0.05$, and $^{***}p<0.01$.}
\label{tab:downstream_main}
\end{table*}

\begin{table*}[!tp]
\centering
\begingroup%
\small%
\setlength{\tabcolsep}{3pt}%
\renewcommand{\arraystretch}{1.05}%
\noindent\begin{tabular*}{\textwidth}{@{\extracolsep{\fill}}lrrrrrr@{}}
\toprule
Model & \multicolumn{3}{c}{Earnings recall (\%)} & \multicolumn{3}{c}{Forecast loss ($\times10^{4}$)} \\
\cmidrule(lr){2-4}\cmidrule(lr){5-7}
& Disc. Pre & Disc. Post & Rel. $\Delta$ (pp) & Disc. Pre & Disc. Post & Rel. $\Delta$ \\
\midrule
Qwen3.8-27B & 20.67 & 13.67 & $-3.00$ & 16.12 & 31.01 & $-3.00$ \\
GPT-5.6-Sol & 87.33 & 31.33 & $-46.67^{***}$ & 13.83 & 30.25 & $-0.76$ \\
GPT-5.6-Luna & 51.67 & 25.33 & $-16.33^{**}$ & 14.22 & 31.79 & $-0.09$ \\
Claude-Haiku-5.5 & 40.33 & 29.67 & $-0.67$ & 16.16 & 31.90 & $-1.92$ \\
\bottomrule
\end{tabular*}\par
\endgroup

\caption{Earnings recall and plain-prompt forecasts on the common 50-company cohort, grouped by result-disclosure month. Disc. Pre/Post denote Disclosure-Pre/Post. Rel. $\Delta$ adjusts for same-question Qwen2.5-72B; loss is displayed $\times10^4$. $^{*}/^{**}/^{***}$ denote raw $p<0.10/0.05/0.01$; accuracy-emphasis loss results and inference are in Appendix Table~\ref{tab:earnings_forecast_supplement}.}
\label{tab:earnings_forecast_main}
\end{table*}

\subsection{Quarterly Earnings}

We use three conditions: earnings recall, plain forecasting, and accuracy-emphasis forecasting. All three require a finite numerical value for diluted earnings per share (EPS) under U.S. Generally Accepted Accounting Principles (GAAP). The recall condition provides only company and quarter information and asks the model to report actual EPS from its existing knowledge. The forecasting conditions cut off available information on the day before the target quarter begins. They provide financial data for the latest four quarters that had been publicly released by that date, EPS for the same quarter of the previous year, and official operating commentary and guidance. Accuracy-emphasis forecasting adds the sentence \emph{Do whatever it takes to give the most accurate answer.} to the plain forecasting instructions. All three conditions require an answer, and the models do not use the web or tools.

A recall response is counted as a hit when its error is no greater than USD 0.05 or 10\% of the absolute actual EPS, whichever is larger, on the original reporting basis. The primary forecasting metric is the absolute EPS error divided by the stock price on the last trading day before the information cutoff; lower values are better. EPS hit rate and year-on-year direction accuracy are supplementary metrics. We first compute each model's Disclosure-Post-minus-Disclosure-Pre change and then subtract the change for Qwen2.5-72B on the same questions to measure the target model's change relative to the control. Company-level net changes are evaluated using two-sided sign-flip tests, with unadjusted $p$ values reported. Complete results for both supplementary metrics are given in Appendix Table~\ref{tab:earnings_forecast_metrics}; sample selection, share-basis adjustments, and scoring details are provided in Appendix~\ref{sec:appendix_downstream}.

\subsection{Results}

The overall ranking results differ by market. The average relative IC across the four models falls by 0.331 in the U.S. market, with an overall one-sided $p=0.0049$. In the Chinese market, it falls by 0.051, with $p=0.2482$. At the model level, all four models show a relative performance decline in the U.S. market. The unadjusted decline tests for Qwen3.8, Sol, and Haiku reach the 5\% level, while Luna's decline is not significant. Results vary more across models in the Chinese market. When the return start is changed to the close of the preceding month's last trading day, the average decline is 0.274 in the U.S. market ($p=0.0006$) and 0.080 in the Chinese market ($p=0.1315$), preserving the market-level conclusion. The appendix reports the full overall and model-level inference results.

Quarterly earnings results are shown in Table~\ref{tab:earnings_forecast_main}. Recall hit rates decline for all four target models. Sol falls from 87.33\% to 31.33\%, and Luna from 51.67\% to 25.33\%. After subtracting the change in the same-question control, their additional declines are 46.67 and 16.33 percentage points, respectively; both unadjusted two-sided tests reach the 5\% level. Qwen3.8 falls from 20.67\% to 13.67\%, and Haiku from 40.33\% to 29.67\%, with neither net change significant.

Forecasting changes across the two result-disclosure periods differ from the recall changes. Under the plain prompt, all four target models show lower EPS hit rates and higher primary forecasting errors, and the same-question 72B control and simple historical baselines also perform worse in Disclosure-Post. Relative to 72B, all eight unadjusted tests of the net change in the primary error metric produce $p$ values above 0.10. Luna's EPS hit rate under the plain prompt falls by an additional 12.33 percentage points relative to the control, with an unadjusted two-sided $p=0.0400$. Year-on-year direction accuracy declines for both the target models and 72B; after subtracting the change in 72B, the relative change is positive for all four models under both prompts. The relative before-and-after differences are concentrated in U.S. asset ranking and earnings recall for Sol and Luna; Chinese-market ranking and primary forecasting error do not show the same pattern.

\section{Conclusion}
\label{sec:conclusion}

Financial knowledge stored in model parameters may extend beyond the time being evaluated, and its temporal scope is difficult to observe directly. FinCutBench measures the temporal scope of a fixed model's knowledge through a continually updated bank of monthly financial fact questions. We form monthly knowledge profiles from closed-book performance over consecutive months, locate Transition Start and FinCut Boundary using a test for a significant decline and piecewise curve fitting, and separately report decline detection, the Boundary point estimate, and localization stability.

The downstream experiments show differences across tasks and markets. Average relative performance across four models declines significantly in U.S. asset ranking, while the Chinese market shows no significant aggregate change. In quarterly earnings tasks, unadjusted two-sided tests show significant declines in earnings recall for Sol and Luna relative to Qwen2.5-72B on the same questions, while the main prediction-error changes paired by result-disclosure period are not significant. FinCut Boundary thus supports evaluation-window selection and cross-period interpretation as a monthly reference for parametric knowledge.

\clearpage
\section{Limitations}
\label{sec:limitations}

FinCutBench covers selected financial facts from the Chinese and U.S. markets. Four-option multiple-choice questions permit some guessing, so observed accuracy may reflect both factual knowledge and the ability to infer answers from the options. The exact time at which each fact first entered the public domain is also difficult to verify individually. We assign question months using verifiable public dates and source-specific rules, with manual corrections for selected samples; earlier professional channels or untraceable dissemination may still make the true first-visible date earlier than the assigned month. For some historical series, we make simplifying assumptions about later revisions. Research report questions are constructed with the help of generative models, and human review covers selected samples. These factors may affect the temporal validity and quality of individual questions.

FinCut Boundary describes overall changes under a given question set and piecewise model, with its scope jointly determined by the market, information axis, question type, and inference configuration. Boundary corresponds to the fitted low level in the later part of the monthly knowledge profile. This level may be above chance, and the model may retain knowledge of individual facts within it. Resampling preserves the question structure within each month, while shared shocks across months and systematic bias from the piecewise model may still affect the date range.

Downstream validation uses fixed pools of Chinese and U.S. assets, a limited number of calendar months, and earnings reports from large U.S. companies. Earnings forecasts are paired by result-disclosure period; a common post-Boundary prediction-decision window for all four models requires later disclosed quarters. Observed differences between periods are jointly influenced by parametric knowledge, market conditions, and task difficulty. Other markets, periods, and tasks require separate validation.

\FloatBarrier
\bibliography{bib/references}

@misc{livebench,
      title={LiveBench: A Challenging, Contamination-Limited {LLM} Benchmark},
      author={Colin White and Samuel Dooley and Manley Roberts and Arka Pal and Ben Feuer and Siddhartha Jain and Ravid Shwartz-Ziv and Neel Jain and Khalid Saifullah and Sreemanti Dey and Shubh-Agrawal and Sandeep Singh Sandha and Siddartha Naidu and Chinmay Hegde and Yann LeCun and Tom Goldstein and Willie Neiswanger and Micah Goldblum},
      year={2025},
      eprint={2406.19314},
      archivePrefix={arXiv},
      primaryClass={cs.CL},
      url={https://arxiv.org/abs/2406.19314},
}

@misc{realtimeqa,
      title={RealTime {QA}: What's the Answer Right Now?},
      author={Jungo Kasai and Keisuke Sakaguchi and Yoichi Takahashi and Ronan Le Bras and Akari Asai and Xinyan Yu and Dragomir Radev and Noah A. Smith and Yejin Choi and Kentaro Inui},
      year={2024},
      eprint={2207.13332},
      archivePrefix={arXiv},
      primaryClass={cs.CL},
      url={https://arxiv.org/abs/2207.13332},
}

@misc{freshqa,
      title={FreshLLMs: Refreshing Large Language Models with Search Engine Augmentation},
      author={Tu Vu and Mohit Iyyer and Xuezhi Wang and Noah Constant and Jerry Wei and Jason Wei and Chris Tar and Yun-Hsuan Sung and Denny Zhou and Quoc Le and Thang Luong},
      year={2023},
      eprint={2310.03214},
      archivePrefix={arXiv},
      primaryClass={cs.CL},
      url={https://arxiv.org/abs/2310.03214},
}

@misc{cdqa,
      title={Let {LLMs} Take on the Latest Challenges! A Chinese Dynamic Question Answering Benchmark},
      author={Zhikun Xu and Yinghui Li and Ruixue Ding and Xinyu Wang and Boli Chen and Yong Jiang and Hai-Tao Zheng and Wenlian Lu and Pengjun Xie and Fei Huang},
      year={2024},
      eprint={2402.19248},
      archivePrefix={arXiv},
      primaryClass={cs.CL},
      url={https://arxiv.org/abs/2402.19248},
}

@misc{dateddata,
      title={Dated Data: Tracing Knowledge Cutoffs in Large Language Models},
      author={Jeffrey Cheng and Marc Marone and Orion Weller and Dawn Lawrie and Daniel Khashabi and Benjamin Van Durme},
      year={2024},
      eprint={2403.12958},
      archivePrefix={arXiv},
      primaryClass={cs.CL},
      url={https://arxiv.org/abs/2403.12958},
}

@misc{llmlagbench,
      title={LLMLagBench: Identifying Temporal Training Boundaries in Large Language Models},
      author={Piotr P\k{e}zik and Konrad Kaczy\'{n}ski and Maria Szyma\'{n}ska and Filip \.{Z}arnecki and Zuzanna Deckert and Jakub Kwiatkowski and Wojciech Janowski},
      year={2025},
      eprint={2511.12116},
      archivePrefix={arXiv},
      primaryClass={cs.CL},
      url={https://arxiv.org/abs/2511.12116},
}

@misc{simulatedignorance,
      title={Simulated Ignorance Fails: A Systematic Study of {LLM} Behaviors on Forecasting Problems Before Model Knowledge Cutoff},
      author={Zehan Li and Yuxuan Wang and Ali El Lahib and Ying-Jieh Xia and Xinyu Pi},
      year={2026},
      eprint={2601.13717},
      archivePrefix={arXiv},
      primaryClass={cs.CL},
      url={https://arxiv.org/abs/2601.13717},
}

@misc{promptsrewind,
      title={Can Prompts Rewind Time for {LLMs}? Evaluating the Effectiveness of Prompted Knowledge Cutoffs},
      author={Xin Gao and Ruiyi Zhang and Daniel Du and Saurabh Mahindre and Sai Ashish Somayajula and Pengtao Xie},
      year={2025},
      eprint={2510.02340},
      archivePrefix={arXiv},
      primaryClass={cs.CL},
      url={https://arxiv.org/abs/2510.02340},
}

@misc{exante,
      title={ExAnte: A Benchmark for Ex-Ante Inference in Large Language Models},
      author={Yachuan Liu and Xiaochun Wei and Lin Shi and Xinnuo Li and Bohan Zhang and Paramveer Dhillon and Qiaozhu Mei},
      year={2025},
      eprint={2505.19533},
      archivePrefix={arXiv},
      primaryClass={cs.LG},
      url={https://arxiv.org/abs/2505.19533},
}

@misc{finqa,
      title={FinQA: A Dataset of Numerical Reasoning over Financial Data},
      author={Zhiyu Chen and Wenhu Chen and Charese Smiley and Sameena Shah and Iana Borova and Dylan Langdon and Reema Moussa and Matt Beane and Ting-Hao Huang and Bryan Routledge and William Yang Wang},
      year={2022},
      eprint={2109.00122},
      archivePrefix={arXiv},
      primaryClass={cs.CL},
      url={https://arxiv.org/abs/2109.00122},
}

@misc{tatqa,
      title={TAT-{QA}: A Question Answering Benchmark on a Hybrid of Tabular and Textual Content in Finance},
      author={Fengbin Zhu and Wenqiang Lei and Youcheng Huang and Chao Wang and Shuo Zhang and Jiancheng Lv and Fuli Feng and Tat-Seng Chua},
      year={2021},
      eprint={2105.07624},
      archivePrefix={arXiv},
      primaryClass={cs.CL},
      url={https://arxiv.org/abs/2105.07624},
}

@misc{convfinqa,
      title={ConvFinQA: Exploring the Chain of Numerical Reasoning in Conversational Finance Question Answering},
      author={Zhiyu Chen and Shiyang Li and Charese Smiley and Zhiqiang Ma and Sameena Shah and William Yang Wang},
      year={2022},
      eprint={2210.03849},
      archivePrefix={arXiv},
      primaryClass={cs.CL},
      url={https://arxiv.org/abs/2210.03849},
}

@misc{financebench,
      title={FinanceBench: A New Benchmark for Financial Question Answering},
      author={Pranab Islam and Anand Kannappan and Douwe Kiela and Rebecca Qian and Nino Scherrer and Bertie Vidgen},
      year={2023},
      eprint={2311.11944},
      archivePrefix={arXiv},
      primaryClass={cs.CL},
      url={https://arxiv.org/abs/2311.11944},
}

@misc{finben,
      title={FinBen: A Holistic Financial Benchmark for Large Language Models},
      author={Qianqian Xie and Weiguang Han and Zhengyu Chen and Ruoyu Xiang and Xiao Zhang and Yueru He and Mengxi Xiao and Dong Li and Yongfu Dai and Duanyu Feng and Yijing Xu and Haoqiang Kang and Ziyan Kuang and Chenhan Yuan and Kailai Yang and Zheheng Luo and Tianlin Zhang and Zhiwei Liu and Guojun Xiong and Zhiyang Deng and Yuechen Jiang and Zhiyuan Yao and Haohang Li and Yangyang Yu and Gang Hu and Jiajia Huang and Xiao-Yang Liu and Alejandro Lopez-Lira and Benyou Wang and Yanzhao Lai and Hao Wang and Min Peng and Sophia Ananiadou and Jimin Huang},
      year={2024},
      eprint={2402.12659},
      archivePrefix={arXiv},
      primaryClass={cs.CL},
      url={https://arxiv.org/abs/2402.12659},
}

@misc{fineval,
      title={FinEval: A Chinese Financial Domain Knowledge Evaluation Benchmark for Large Language Models},
      author={Xin Guo and Haotian Xia and Zhaowei Liu and Hanyang Cao and Zhi Yang and Zhiqiang Liu and Sizhe Wang and Jinyi Niu and Chuqi Wang and Yanhui Wang and Xiaolong Liang and Xiaoming Huang and Bing Zhu and Zhongyu Wei and Yun Chen and Weining Shen and Liwen Zhang},
      year={2024},
      eprint={2308.09975},
      archivePrefix={arXiv},
      primaryClass={cs.CL},
      url={https://arxiv.org/abs/2308.09975},
}

@inproceedings{investorbench,
    title = "{INVESTORBENCH}: A Benchmark for Financial Decision-Making Tasks with {LLM}-based Agent",
    author = "Li, Haohang  and
      Cao, Yupeng  and
      Yu, Yangyang  and
      Javaji, Shashidhar Reddy  and
      Deng, Zhiyang  and
      He, Yueru  and
      Jiang, Yuechen  and
      Zhu, Zining  and
      Subbalakshmi, K.P.  and
      Huang, Jimin  and
      Qian, Lingfei  and
      Peng, Xueqing  and
      Suchow, Jordan W.  and
      Xie, Qianqian",
    editor = "Che, Wanxiang  and
      Nabende, Joyce  and
      Shutova, Ekaterina  and
      Pilehvar, Mohammad Taher",
    booktitle = "Proceedings of the 63rd Annual Meeting of the {A}ssociation for {C}omputational {L}inguistics (Volume 1: Long Papers)",
    month = jul,
    year = "2025",
    address = "Vienna, Austria",
    publisher = "Association for Computational Linguistics",
    url = "https://aclanthology.org/2025.acl-long.126/",
    doi = "10.18653/v1/2025.acl-long.126",
    pages = "2509--2525",
    ISBN = "979-8-89176-251-0"
}

@misc{fintradebench,
      title={FinTradeBench: A Financial Reasoning Benchmark for LLMs},
      author={Yogesh Agrawal and Aniruddha Dutta and Md Mahadi Hasan and Santu Karmaker and Aritra Dutta},
      year={2026},
      eprint={2603.19225},
      archivePrefix={arXiv},
      primaryClass={cs.CE},
      url={https://arxiv.org/abs/2603.19225},
}

@inproceedings{historicalfinanceqa,
  title = {Beyond the Reported Cutoff: Where Large Language Models Fall Short on Financial Knowledge},
  author = {Agam Shah and Liqin Ye and Sebastian Jaskowski and Wei Xu and Sudheer Chava},
  booktitle = {COLM 2025},
  year = {2025},
  url = {https://openreview.net/forum?id=lEpPFmGH3L},
}

@misc{lookaheadbench,
      title={Look-Ahead-Bench: a Standardized Benchmark of Look-ahead Bias in Point-in-Time {LLMs} for Finance},
      author={Mostapha Benhenda},
      year={2026},
      eprint={2601.13770},
      archivePrefix={arXiv},
      primaryClass={cs.AI},
      url={https://arxiv.org/abs/2601.13770},
}

@misc{profitmirage,
      title={Profit Mirage: Revisiting Information Leakage in {LLM}-based Financial Agents},
      author={Xiangyu Li and Yawen Zeng and Xiaofen Xing and Jin Xu and Xiangmin Xu},
      year={2025},
      eprint={2510.07920},
      archivePrefix={arXiv},
      primaryClass={cs.AI},
      url={https://arxiv.org/abs/2510.07920},
}

@misc{datedgpt,
      title={DatedGPT: Preventing Lookahead Bias in Large Language Models with Time-Aware Pretraining},
      author={Yutong Yan and Raphael Tang and Zhenyu Gao and Wenxi Jiang and Yao Lu},
      year={2026},
      eprint={2603.11838},
      archivePrefix={arXiv},
      primaryClass={cs.CL},
      url={https://arxiv.org/abs/2603.11838},
}

@misc{goalmissingtruth,
      title={Seeing the Goal, Missing the Truth: Human Accountability for {AI} Bias},
      author={Sean Cao and Wei Jiang and Hui Xu},
      year={2026},
      eprint={2602.09504},
      archivePrefix={arXiv},
      primaryClass={q-fin.GN},
      url={https://arxiv.org/abs/2602.09504},
}

@misc{llmforecaststocks,
      title={Can {ChatGPT} Forecast Stock Price Movements? Return Predictability and Large Language Models},
      author={Alejandro Lopez-Lira and Yuehua Tang},
      year={2025},
      eprint={2304.07619},
      archivePrefix={arXiv},
      primaryClass={q-fin.ST},
      url={https://arxiv.org/abs/2304.07619},
}

@misc{glassermanlookahead,
      title={Assessing Look-Ahead Bias in Stock Return Predictions Generated By {GPT} Sentiment Analysis},
      author={Paul Glasserman and Caden Lin},
      year={2023},
      eprint={2309.17322},
      archivePrefix={arXiv},
      primaryClass={q-fin.GN},
      url={https://arxiv.org/abs/2309.17322},
}

@misc{memorizationforecast,
      title={The Memorization Problem: Can We Trust {LLMs}' Economic Forecasts?},
      author={Alejandro Lopez-Lira and Yuehua Tang and Mingyin Zhu},
      year={2025},
      eprint={2504.14765},
      archivePrefix={arXiv},
      primaryClass={q-fin.GN},
      url={https://arxiv.org/abs/2504.14765},
}

@misc{stockbench,
      title={StockBench: Can {LLM} Agents Trade Stocks Profitably In Real-world Markets?},
      author={Yanxu Chen and Zijun Yao and Yantao Liu and Amy Xin and Jin Ye and Jianing Yu and Lei Hou and Juanzi Li},
      year={2026},
      eprint={2510.02209},
      archivePrefix={arXiv},
      primaryClass={cs.LG},
      url={https://arxiv.org/abs/2510.02209},
}

@misc{qwen25,
  title = {{Qwen2.5 Technical Report}},
  author = {{Qwen Team}},
  year = {2025},
  url = {https://arxiv.org/abs/2412.15115},
  note = {arXiv:2412.15115.},
}

@inproceedings{fintextqa,
    title = "{F}in{T}ext{QA}: A Dataset for Long-form Financial Question Answering",
    author = "Chen, Jian  and
      Zhou, Peilin  and
      Hua, Yining  and
      Xin, Loh  and
      Chen, Kehui  and
      Li, Ziyuan  and
      Zhu, Bing  and
      Liang, Junwei",
    editor = "Ku, Lun-Wei  and
      Martins, Andre  and
      Srikumar, Vivek",
    booktitle = "Proceedings of the 62nd Annual Meeting of the {A}ssociation for {C}omputational {L}inguistics (Volume 1: Long Papers)",
    month = aug,
    year = "2024",
    address = "Bangkok, Thailand",
    publisher = "Association for Computational Linguistics",
    url = "https://aclanthology.org/2024.acl-long.328/",
    doi = "10.18653/v1/2024.acl-long.328",
    pages = "6025--6047"
}

@misc{hindsightbench,
      title={HindsightBench: A Black-Box Behavioral Audit Protocol for Parametric Hindsight in Time-Indexed {LLM} Decision Tasks},
      author={Haozhe Jia},
      year={2026},
      eprint={2607.18867},
      archivePrefix={arXiv},
      primaryClass={cs.LG},
      url={https://arxiv.org/abs/2607.18867},
}

@misc{knowmtbench,
      title={KnowMT-Bench: Benchmarking Knowledge-Intensive Long-Form Question Answering in Multi-Turn Dialogues},
      author={Junhao Chen and Yu Huang and Siyuan Li and Rui Yao and Hanqian Li and Hanyu Zhang and Jungang Li and Jian Chen and Bowen Wang and Xuming Hu},
      year={2026},
      eprint={2509.21856},
      archivePrefix={arXiv},
      primaryClass={cs.CL},
      url={https://arxiv.org/abs/2509.21856},
}

@misc{chroknowledge,
      title={ChroKnowledge: Unveiling Chronological Knowledge of Language Models in Multiple Domains},
      author={Yein Park and Chanwoong Yoon and Jungwoo Park and Donghyeon Lee and Minbyul Jeong and Jaewoo Kang},
      year={2025},
      eprint={2410.09870},
      archivePrefix={arXiv},
      primaryClass={cs.CL},
      url={https://arxiv.org/abs/2410.09870},
}

@misc{detectinglookahead,
      title={Detecting Lookahead Bias in {LLM} Forecasts},
      author={Zhenyu Gao and Wenxi Jiang and Yutong Yan},
      year={2026},
      eprint={2512.23847},
      archivePrefix={arXiv},
      primaryClass={q-fin.GN},
      url={https://arxiv.org/abs/2512.23847},
}

@misc{scalingpit,
      title={Scaling Point-in-Time Language Models: Enabling Economic Evaluation of Embeddings},
      author={Bryan Kelly and Semyon Malamud and Johannes Schwab and Teng Andrea Xu},
      year={2026},
      eprint={2607.11889},
      archivePrefix={arXiv},
      primaryClass={cs.CL},
      url={https://arxiv.org/abs/2607.11889},
}

@misc{fincut_deepseek_v41,
  title = {{DeepSeek-V4.1-Flash-1017 Release}},
  author = {{DeepSeek-AI}},
  year = {2026},
  url = {https://api-docs.deepseek.com/news/news261017/},
  note = {Official release announcement for the evaluated 1017 version.},
  urldate = {2026-11-09},
}

@misc{fincut_gpt56_sol,
  title = {{GPT-5.6 System Card}},
  author = {{OpenAI}},
  year = {2026},
  url = {https://deploymentsafety.openai.com/gpt-5-6},
  note = {Official system card; published 2026-07-09.},
  urldate = {2026-10-24},
}

@misc{fincut_gpt61_sol,
  title = {{Addendum to {GPT}-6 Astra System Card: {GPT}-6.1 Sol}},
  author = {{OpenAI}},
  year = {2026},
  url = {https://deploymentsafety.openai.com/gpt-6-1-sol},
  note = {Official system-card addendum; published 2026-09-29.},
  urldate = {2026-10-24},
}

@misc{fincut_gpt6_astra,
  title = {{GPT-6 Astra System Card}},
  author = {{OpenAI}},
  year = {2026},
  url = {https://deploymentsafety.openai.com/gpt-6-astra},
  note = {Official system card; published 2026-09-03.},
  urldate = {2026-10-24},
}

@misc{fincut_qwen35,
    title  = {{Qwen3.5}: Towards Native Multimodal Agents},
    author = {{Qwen Team}},
    month  = {February},
    year   = {2026},
    url    = {https://qwen.ai/blog?id=qwen3.5}
}

@misc{fincut_qwen38,
  title = {{Qwen3.8-27B Model Card}},
  author = {{Qwen Team}},
  year = {2026},
  url = {https://huggingface.co/Qwen/Qwen3.8-27B},
  note = {Official model card for the evaluated 27B checkpoint.},
  urldate = {2026-11-09},
}

@misc{googledeepmind2026gemini38flashcard,
  title = {{Gemini 3.8 Flash Model Card}},
  author = {{Google DeepMind}},
  year = {2026},
  url = {https://deepmind.google/models/model-cards/gemini-3-8-flash/},
  note = {Official model card; published 2026-09-02.},
  urldate = {2026-10-24},
}

@misc{fincut_gpt56_sol_api,
  title = {{GPT-5.6 Sol Model}},
  author = {{OpenAI}},
  year = {2026},
  url = {https://developers.openai.com/api/docs/models/gpt-5.6-sol},
  note = {Official API model documentation.},
  urldate = {2026-10-26},
}

@misc{fincut_gpt6_astra_api,
  title = {{GPT-6 Astra Model}},
  author = {{OpenAI}},
  year = {2026},
  url = {https://developers.openai.com/api/docs/models/gpt-6-astra},
  note = {Official API model documentation.},
  urldate = {2026-10-26},
}

@misc{fincut_cut_gpt6_astra_training_summary,
  title = {{Public Summary of Training Content for {GPT}-6 Astra}},
  author = {{OpenAI}},
  year = {2026},
  url = {https://cdn.openai.com/pdf/gpt-6-astra-eu-ai-act-public-summary-of-training-content.pdf},
  note = {Version 1; updated 2026-09-02.},
  urldate = {2026-10-31},
}

@misc{fincut_deepseek_v41_training_summary,
  title = {{Public Summary of Training Content for DeepSeek-V4.1}},
  author = {{DeepSeek}},
  year = {2026},
  url = {https://cdn.deepseek.com/policies/DeepSeek-V4.1-public-summary-of-training-content.pdf},
  note = {Updated 2026-09-24.},
  urldate = {2026-10-31},
}

@article{finmem2025, title={FinMem: A Performance-Enhanced {LLM} Trading Agent With Layered Memory and Character Design}, volume={11}, ISSN={2372-2096}, url={http://dx.doi.org/10.1109/TBDATA.2025.3593370}, DOI={10.1109/tbdata.2025.3593370}, number={6}, journal={IEEE Transactions on Big Data}, publisher={Institute of Electrical and Electronics Engineers (IEEE)}, author={Yu, Yangyang and Li, Haohang and Chen, Zhi and Jiang, Yuechen and Li, Yang and Suchow, Jordan W. and Zhang, Denghui and Khashanah, Khaldoun}, year={2025}, month=Dec, pages={3443–3459} }

@inproceedings{finagent2024, series={KDD ’24}, title={A Multimodal Foundation Agent for Financial Trading: Tool-Augmented, Diversified, and Generalist}, url={http://dx.doi.org/10.1145/3637528.3671801}, DOI={10.1145/3637528.3671801}, booktitle={Proceedings of the 30th ACM SIGKDD Conference on Knowledge Discovery and Data Mining}, publisher={ACM}, author={Zhang, Wentao and Zhao, Lingxuan and Xia, Haochong and Sun, Shuo and Sun, Jiaze and Qin, Molei and Li, Xinyi and Zhao, Yuqing and Zhao, Yilei and Cai, Xinyu and Zheng, Longtao and Wang, Xinrun and An, Bo}, year={2024}, month=Aug, pages={4314–4325}, collection={KDD ’24} }

@inproceedings{fincon2024,
 author = {Yu, Yangyang and Yao, Zhiyuan and Li, Haohang and Deng, Zhiyang and Jiang, Yuechen and Cao, Yupeng and Chen, Zhi and Suchow, Jordan W. and Cui, Zhenyu and Liu, Rong and Xu, Zhaozhuo and Zhang, Denghui and Subbalakshmi, Koduvayur and Xiong, Guojun and He, Yueru and Huang, Jimin and Li, Dong and Xie, Qianqian},
 booktitle = {Advances in Neural Information Processing Systems},
 doi = {10.52202/079017-4354},
 editor = {A. Globerson and L. Mackey and D. Belgrave and A. Fan and U. Paquet and J. Tomczak and C. Zhang},
 pages = {137010--137045},
 publisher = {Curran Associates, Inc.},
 title = {FinCon: A Synthesized {LLM} Multi-Agent System with Conceptual Verbal Reinforcement for Enhanced Financial Decision Making},
 url = {https://proceedings.neurips.cc/paper_files/paper/2024/file/f7ae4fe91d96f50abc2211f09b6a7e49-Paper-Conference.pdf},
 volume = {37},
 year = {2024}
}

@misc{tradingagents2025,
      title={TradingAgents: Multi-Agents {LLM} Financial Trading Framework},
      author={Yijia Xiao and Edward Sun and Di Luo and Wei Wang},
      year={2025},
      eprint={2412.20138},
      archivePrefix={arXiv},
      primaryClass={q-fin.TR},
      url={https://arxiv.org/abs/2412.20138},
}

@inproceedings{finsaber2026, title={Can {LLM}-based Financial Investing Strategies Outperform the Market in Long Run?}, url={http://dx.doi.org/10.1145/3770854.3785702}, DOI={10.1145/3770854.3785702}, booktitle={Proceedings of the 32nd ACM SIGKDD Conference on Knowledge Discovery and Data Mining V.1}, publisher={ACM}, author={Li, Weixian Waylon and Kim, Hyeonjun and Cucuringu, Mihai and Ma, Tiejun}, year={2026}, month=Apr, pages={2711–2722} }

@inproceedings{deepfund2025,
  title = {{Time Travel is Cheating: Going Live with DeepFund for Real-Time Fund Investment Benchmarking}},
  author = {Changlun Li and Yao Shi and Chen Wang and Qiqi Duan and Runke Ruan and Weijie Huang and Haonan Long and Lijun Huang and Nan Tang and Yuyu Luo},
  booktitle = {Advances in Neural Information Processing Systems},
  year = {2025},
  url = {https://doi.org/10.52202/085713-5554},
  doi = {10.52202/085713-5554},
}

@misc{ktdfin2026,
      title={From Knowing to Doing: A Memory-Controlled Benchmark for {LLM} Trading Agents on Stock Markets},
      author={Taojie Zhu and Wentao Zhao and Rui Sun and Beidi Luan and Jiacheng Lu and Sinuo Wang and Jing Li and Daxin Jiang and Yonghong He and Zuo Bai},
      year={2026},
      eprint={2605.28359},
      archivePrefix={arXiv},
      primaryClass={cs.AI},
      url={https://arxiv.org/abs/2605.28359},
}

@misc{fincut_claude_haiku45_api,
  title = {{Claude Haiku 4.5}},
  author = {{Anthropic}},
  year = {2025},
  url = {https://platform.claude.com/docs/en/models/haiku-4-5/overview},
  note = {Official model documentation; released October 15, 2025.},
  urldate = {2026-10-31},
}

@misc{fincut_claude_haiku55_api,
  title = {{Claude Haiku 5.5}},
  author = {{Anthropic}},
  year = {2026},
  url = {https://platform.claude.com/docs/en/models/haiku-5-5/overview},
  note = {Official model documentation; released October 7, 2026.},
  urldate = {2026-10-31},
}

@article{gu2020empirical,
  author = {Gu, Shihao and Kelly, Bryan and Xiu, Dacheng},
  title = {Empirical Asset Pricing via Machine Learning},
  journal = {The Review of Financial Studies},
  year = {2020},
  volume = {33},
  number = {5},
  pages = {2223--2273},
  doi = {10.1093/rfs/hhaa009},
  url = {https://academic.oup.com/rfs/article/33/5/2223/5758276}
}

@inproceedings{qin-etal-2024-large,
    title = "Large Language Models are Effective Text Rankers with Pairwise Ranking Prompting",
    author = "Qin, Zhen  and
      Jagerman, Rolf  and
      Hui, Kai  and
      Zhuang, Honglei  and
      Wu, Junru  and
      Yan, Le  and
      Shen, Jiaming  and
      Liu, Tianqi  and
      Liu, Jialu  and
      Metzler, Donald  and
      Wang, Xuanhui  and
      Bendersky, Michael",
    editor = "Duh, Kevin  and
      Gomez, Helena  and
      Bethard, Steven",
    booktitle = "Findings of the {A}ssociation for {C}omputational {L}inguistics: {NAACL} 2024",
    month = jun,
    year = "2024",
    address = "Mexico City, Mexico",
    publisher = "Association for Computational Linguistics",
    url = "https://aclanthology.org/2024.findings-naacl.97/",
    doi = "10.18653/v1/2024.findings-naacl.97",
    pages = "1504--1518"
}

@inproceedings{sun-etal-2023-chatgpt,
    title = "Is {C}hat{GPT} Good at Search? Investigating Large Language Models as Re-Ranking Agents",
    author = "Sun, Weiwei  and
      Yan, Lingyong  and
      Ma, Xinyu  and
      Wang, Shuaiqiang  and
      Ren, Pengjie  and
      Chen, Zhumin  and
      Yin, Dawei  and
      Ren, Zhaochun",
    editor = "Bouamor, Houda  and
      Pino, Juan  and
      Bali, Kalika",
    booktitle = "Proceedings of the 2023 Conference on Empirical Methods in Natural Language Processing",
    month = dec,
    year = "2023",
    address = "Singapore",
    publisher = "Association for Computational Linguistics",
    url = "https://aclanthology.org/2023.emnlp-main.923/",
    doi = "10.18653/v1/2023.emnlp-main.923",
    pages = "14918--14937"
}

@inproceedings{fincad2026,
    title = "Summoning the Oracle to Slay It: Mitigating Look-Ahead Bias in Financial Backtesting with Large Language Models",
    author = "Li, Weixian Waylon  and
      Wang, Mengyu  and
      Ma, Tiejun",
    editor = "Lee, Hung-Yi  and
      Dev, Sunipa  and
      Li, Jessy  and
      Elliott, Desmond",
    booktitle = "Proceedings of the 2026 Conference on Empirical Methods in Natural Language Processing",
    month = oct,
    year = "2026",
    address = "Budapest, Hungary",
    publisher = "Association for Computational Linguistics",
    url = "https://aclanthology.org/2026.emnlp-main.510/",
    doi = "10.18653/2026.emnlp-main.510",
    pages = "11694--11713",
    ISBN = "979-8-89176-459-0"
}

@article{cameron2008bootstrap, title={Bootstrap-Based Improvements for Inference with Clustered Errors}, volume={90}, ISSN={1530-9142}, url={http://dx.doi.org/10.1162/rest.90.3.414}, DOI={10.1162/rest.90.3.414}, number={3}, journal={Review of Economics and Statistics}, publisher={MIT Press}, author={Cameron, A. Colin and Gelbach, Jonah B. and Miller, Douglas L.}, year={2008}, month=Aug, pages={414–427} }

@misc{fincut_deepseek_v41_report,
  title = {{DeepSeek-V4.1-Flash: Pushing the Limits of KV Cache Compression}},
  author = {{DeepSeek-AI}},
  year = {2026},
  url = {https://huggingface.co/deepseek-ai/DeepSeek-V4.1-Flash/blob/main/DeepSeek\_V41\_Tech\_Report.pdf},
  note = {Official technical report.},
  urldate = {2026-11-09},
}
\bibliographystyle{iclr2027_conference}

\clearpage
\appendix
\section{Dataset Details}
\label{sec:appendix_benchmark_details}

This section describes the data sources, monthly assignment rules, question inclusion criteria, and human review, and presents the dataset coverage and representative questions.

\subsection{Project Repository}
\label{sec:data_availability}

The project repository at \url{https://github.com/FinCutBench/FinCutBench} is the canonical location for information about the monthly question sets, construction and evaluation code, model measurement results, and project updates. Third-party source materials remain subject to their original licenses.

\subsection{Temporal Assignment}

\begin{figure}[H]
\centering
\includegraphics[width=\textwidth]{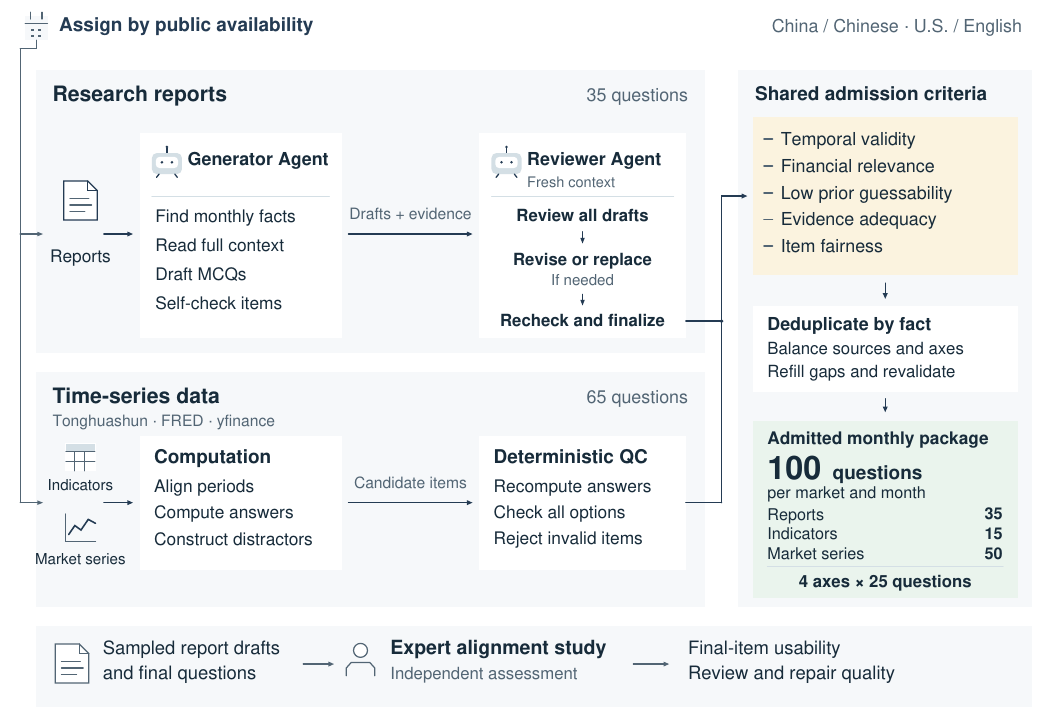}
\caption{\textbf{FinCutBench construction pipeline.} Research reports and time series yield monthly factual multiple-choice questions through source-specific construction, shared admission criteria, independent checks, and balanced packaging.}
\label{fig:fincutbench_construction}
\end{figure}

FinCutBench questions are constructed from financial time series and securities research reports. Chinese time series come from Tonghuashun, U.S. structured indicators come from FRED, and U.S. market data are retrieved through yfinance. Event questions are built from securities research reports purchased through third-party providers. The reports are used to identify candidate facts, construct questions, and verify answers. Questions retain only the entities, times, actions, values, and states directly supported by the report text; broker views, ratings, investment recommendations, forecasts, and report titles are not used as answers to factual event questions. For structured indicators and market time series, the observation window, units, comparison method, and correct answer are determined programmatically. An independent calculation then checks whether each question is consistent with the source data.

Both the generation and quality checks of report-based questions use GPT-6.1-Sol~\cite{fincut_gpt61_sol} through Codex with the high reasoning setting. Quality checks use an independent context to recheck each question against its evidence.

A question is assigned to a month based on when its information became public. For time series without exact release dates or historical versions, source-specific rules determine when the data can be used. The statistical reference period identifies the period described by an observation, the disclosure time identifies when the answer became publicly available, and the reference date in the question specifies the historical state on which the model should base its answer. These three times may differ and must be stated separately. U.S. structured indicators use historical versions of FRED data available at the end of the assigned month. When Chinese structured indicators lack exact release dates and historical versions, we apply a simplified conservative rule: the question set for month $t$ uses only monthly observations from $t-1$ or earlier, and treats the $t-1$ observation as available by the end of month $t$. The latest data version available at collection is used only for series with a low risk of later revisions and verified indicator definitions. Report-based events are assigned to the relevant window according to the public disclosure time of the tested fact. Their status at that time is verified using identifiable materials produced within the target window, without using later materials to backfill earlier windows.

Market-data questions are assigned to the relevant window once the full price interval is available, and closing data are treated as available after the market closes that day. For Chinese stocks and ETFs, we use a continuous adjusted-price index constructed by chaining the vendor's daily percentage changes; for Chinese indices, we use unadjusted closing index levels. U.S. assets use the vendor's automatically adjusted prices. Market paths and individual asset moves describe the historical performance of broad markets and specific assets, respectively. They use the same time boundaries, are assigned to their respective information axes, and are reported separately.

\subsection{Time-Series Coverage}
\label{sec:appendix_timeseries_coverage}

Structured indicators cover macroeconomic conditions, money and credit, interest rates, sector demand, upstream prices, inventories, and production capacity. Table~\ref{tab:appendix_structured_indicators} summarizes the main coverage. Low-frequency data such as GDP and their official releases are covered by dynamically constructed questions. The table lists indicator series that can consistently support numerical questions at monthly or higher frequencies. Indicator questions use observations available within the target window and test specific values, month-on-month differences, or year-on-year differences.

\begin{table}[H]
\centering
\footnotesize
\resizebox{\textwidth}{!}{
\begingroup
\renewcommand{\arraystretch}{1.18}
\setlength{\tabcolsep}{5pt}
\begin{tabular}{@{}p{1.2cm}p{2.2cm}p{11.85cm}@{}}
\toprule
Market & Group & Covered Series \\
\midrule
\multirow{2}{*}{\textbf{CN}} & Macro & Manufacturing PMI; Manufacturing PMI New Orders; CPI YoY; PPI YoY; M2 YoY and level; Total Social Financing Stock YoY; Exports YoY; Industrial Value Added YoY; Retail Sales YoY; Real-Estate Development Investment YoY; Commercial-Building Sales Area YoY; Manufacturing Fixed-Asset Investment YoY. \\
 & Sector & Auto Sales; New-Energy Vehicle Sales and Output; Aluminum and LME Copper Inventories; Polysilicon, A00 Aluminum Ingot, Lithium Carbonate, Rebar, and Thermal-Coal Prices; Road Logistics and Coastal Container Freight Indices. \\
\addlinespace[2pt]
\multirow{2}{*}{\textbf{US}} & Macro & CPI and Core CPI; PCE and Core PCE Price Indices; Unemployment Rate; Nonfarm Payroll Employment; Average Hourly Earnings; Retail Sales; Industrial Production; Capacity Utilization; M2 Money Stock; Federal Funds Rate; 3-Month Treasury Bill Rate; 2-, 5-, and 10-Year Treasury Yields; Aaa and Baa Corporate Bond Yields; 10-Year--2-Year Treasury Yield Spread. \\
 & Sector & WTI and Brent Crude-Oil Prices; Henry Hub Natural-Gas Price; Total Vehicle and Heavy-Truck Sales; Housing Starts; Building Permits; Durable-Goods Orders; Manufacturers' New Orders. \\
\bottomrule
\end{tabular}
\endgroup
}
\caption{Structured economic and sector indicators covered by FinCutBench.}
\label{tab:appendix_structured_indicators}
\end{table}

Market data cover core indices, broad-market and sector ETFs, commodity and bond ETFs, and representative Chinese and U.S. stocks. Table~\ref{tab:appendix_market_series} presents representative coverage. Indices and ETFs mainly describe market paths and risk factors, while stocks mainly describe individual asset moves. The questions selected each month depend on data completeness within the target window, suitability for question construction, and whether the questions are too concentrated on particular topics.

\begin{table*}[!tp]
\centering
\footnotesize
\resizebox{\textwidth}{!}{
\begingroup
\renewcommand{\arraystretch}{1.18}
\setlength{\tabcolsep}{5pt}
\begin{tabular}{@{}p{1.2cm}p{2.2cm}p{11.85cm}@{}}
\toprule
Market & Group & Covered Assets \\
\midrule
\multirow{3}{*}{\textbf{CN}} & Indices & CSI 300; SSE Composite; Shenzhen Component; ChiNext; CSI 500; CSI 1000; STAR 50; SSE 50; CSI 800. \\
 & ETFs & CSI 300; CSI 500; ChiNext; SSE 50; STAR 50; CSI 1000; Semiconductor; Securities; Photovoltaic; New-Energy Vehicle; Dividend; Dividend Low-Volatility; Gold; Treasury Bond; 10-Year Treasury Bond; Convertible Bond; Hang Seng; Hang Seng TECH; Hong Kong Connect Internet; China Internet; Nasdaq; S\&P 500. \\
 & Stocks & Kweichow Moutai; CATL; China Ping An; China Merchants Bank; Wuliangye Yibin; Midea Group; BYD; Hikvision; Luxshare Precision; Zijin Mining; CITIC Securities; Hengrui Medicine; China Yangtze Power; China Tourism Group Duty Free; WuXi AppTec; SMIC; Hygon Information Technology; LONGi Green Energy; East Money Information; Mindray Bio-Medical Electronics; iFLYTEK; COSCO Shipping Holdings. \\
\addlinespace[2pt]
\multirow{3}{*}{\textbf{US}} & Indices & S\&P 500; Nasdaq Composite; Nasdaq 100; Dow Jones Industrial Average; Russell 2000; VIX. \\
 & ETFs & SPY; QQQ; IWM; DIA; GLD; TLT; HYG; USO; XLK; XLF; XLE; SMH. \\
 & Stocks & Apple; Microsoft; Nvidia; Amazon; Meta Platforms; Alphabet; Tesla; JPMorgan Chase; Berkshire Hathaway; Exxon Mobil; UnitedHealth Group; Broadcom; Costco Wholesale; Eli Lilly; AMD. \\
\bottomrule
\end{tabular}
\endgroup
}
\caption{Market indices, exchange-traded funds, and representative stocks covered by FinCutBench.}
\label{tab:appendix_market_series}
\end{table*}

For each market, we select a fixed core set of 15 series covering macroeconomic conditions, sector cycles, and major risk factors: six macro indicators, three sector indicators, and six core market assets. When valid observations are unavailable, a substitute is selected within the same information axis in a fixed order. The remaining indicators and assets are selected based on that month's data completeness and information value. The fixed core set and substitution records form part of the question-set specification.

\subsection{Question Inclusion Criteria}
\label{sec:appendix_question_criteria}

The review uses the following five criteria.

\textbf{Temporal validity.} We check the statistical reference period, information disclosure time, and reference date in the question separately. Report-based questions test facts newly disclosed within the target window, while the months and data versions of structured indicators follow the source-specific rules above. Relative time expressions are tied to an explicit date or a clearly identified version of an event.

\textbf{Financial relevance.} Facts must concern policy, sector, company, or market information relevant to financial research and have clear information value. Subjective ratings, investment recommendations, and report titles are not used as answers to factual questions.

\textbf{Dependence on new information.} Assuming that the question is clearly worded, we remove the new information from the target window and check whether prior knowledge, fixed calendars, institutional rules, scheduled plans, or long-term trends are sufficient to infer the answer. Only facts that require new information are included.

\textbf{Evidence sufficiency.} Credible and identifiable source text or data must directly support the tested fact, the answer, and any necessary conditions. We obtain additional evidence when missing definitions or conflicting materials affect the answer, and exclude questions when key evidence is insufficient.

\textbf{Question wording and option design.} The question must specify all necessary conditions and have a unique answer. The four options must be consistent in type, units, granularity, and tone, with plausible distractors in the same financial context. For numerical questions, we check the calculation basis, interval, and precision. For text-based questions, we check the entities, actions, states, and times. We also check for clues from wording, formatting, or numerical scale that could allow the answer to be guessed.

\subsection{Representative Examples}

Table~\ref{tab:appendix_representative_examples} presents representative questions from the August 2024 question set across all four axes, both languages, and three source types, including the questions, options, answer keys, and evidence summaries. Appendix~\ref{sec:data_availability} identifies the canonical repository for project resources and updates.

\begin{table*}[!tp]
\centering
\small
\resizebox{\textwidth}{!}{
\begingroup
\renewcommand{\arraystretch}{1.15}
\setlength{\tabcolsep}{4pt}
\begin{tabular}{@{}p{1.8cm}p{8.4cm}p{5.0cm}@{}}
\toprule
Axis / market & Question and options & Answer and evidence summary \\
\midrule
Macro / policy\newline US & From 2023-07 to 2024-07, what was the year-over-year percentage change in U.S. CPI for all urban consumers (seasonally adjusted)?\newline A. 2.2\%\newline B. 2.0\%\newline C. 2.6\%\newline D. 2.9\% & \textbf{D}. Dividing the 2024-07 observation by the 2023-07 observation and subtracting 1 gives 2.9\%. \\
Sector / supply\newline CN & According to China Passenger Car Association data released in August 2024, what were domestic retail sales of new energy passenger vehicles in July 2024?\newline A. 838,000 vehicles\newline B. 878,000 vehicles\newline C. 798,000 vehicles\newline D. 758,000 vehicles & \textbf{B}. The association reported July retail sales of 878,000 new energy passenger vehicles. This figure uses the monthly retail-sales measure, which differs from wholesale, export, and cumulative sales measures. \\
Market path\newline US & In 2024-08, on which trading day did the S\&P 500 reach its lowest index closing value?\newline A. 2024-08-08\newline B. 2024-08-06\newline C. 2024-08-05\newline D. 2024-08-07 & \textbf{C}. Comparing daily index closing values within the target month, the lowest value was 5186.3301 on 2024-08-05. \\
Asset moves\newline CN & Based on adjusted closing prices, which option is closest to the cumulative return of COSCO SHIPPING Holdings (601919.SH) from the close on 2024-08-01 to the close on 2024-08-30?\newline A. 0.3\%\newline B. -2.1\%\newline C. -16.9\%\newline D. -5.6\% & \textbf{D}. Recalculating the cumulative return from adjusted closing prices between the close on 2024-08-01 and the close on 2024-08-30 gives -5.58\%. \\
\bottomrule
\end{tabular}
\endgroup
}
\caption{Representative questions from the August 2024 package, with options, answer keys, and evidence summaries. Chinese questions are translated into English.}
\label{tab:appendix_representative_examples}
\end{table*}

\subsection{Human Alignment Protocol}
\label{sec:appendix_human_alignment_protocol}

Human review separately evaluates whether final report-based questions are directly usable and how well automated quality checks work. We sample final questions from the Chinese and English questions for January 2023, March 2025, and September 2026 that were prioritized for release, stratifying by month, market, and information axis. We then sample the initial drafts of some of these questions for paired comparisons. Versions with identical content share one judgment, while versions with different content are reviewed separately. Three experts independently review every distinct version.

Experts first read the question and options, then check the answer and evidence, and make a pass/fail judgment for each of the five criteria. Questions are presented in random order with neutral identifiers; initial--final draft links, quality-check comments, and generation settings are hidden. An expert judges a question directly usable only if it passes all five criteria; insufficient evidence is counted as a failure. Agreement by at least two experts determines the question-level majority judgment. We also calculate the unanimous agreement rate across the three experts, pairwise observed agreement, and Fleiss' $\kappa$.

For final questions, we report the proportion judged directly usable by a majority and its Wilson 95\% confidence interval. Paired initial and final drafts are used to compare the issues flagged in the first check and the effects of revisions. Detection and repair success rates are estimated only when the sample includes initial drafts with problems confirmed by the human majority. When multiple generation models are compared, the proportion of directly usable questions is calculated separately for each model.

\subsection{Human Alignment Results}
\label{sec:appendix_human_alignment_results}

The three experts reviewed \HumanUniqueN{} unique question versions generated by GPT-6.1-Sol-High, covering \HumanFinalN{} final questions and \HumanInitialN{} initial--final draft pairs. Table~\ref{tab:human_alignment_breakdown} summarizes the main results: all final questions passed the majority review, with a Wilson 95\% confidence interval of \HumanWilsonLo{}--\HumanWilsonHi{}. On \HumanDisagreementN{} questions, one of the three reviewers disagreed. These disagreements concerned entity wording, evidence alignment, and financial information value. The large majority of review judgments were passes, producing highly imbalanced category margins. Fleiss' $\kappa=\HumanKappa{}$ is sensitive to these margins; we therefore also report unanimous agreement and pairwise observed agreement.

During quality control, \HumanChangedN{} questions were revised: \HumanStemChangedN{} had reporting dates added, and \HumanOptionsChangedN{} had numerical distractors adjusted. The date-related question flagged in the first check was accepted by all three experts both before and after revision, showing that the automated quality check applied a stricter time constraint than the human majority in this case. All initial drafts passed the majority review; estimating problem-detection recall and repair success therefore requires a sample containing human-confirmed failures.

\begin{table}[!htbp]
\centering\small
\begingroup
\renewcommand{\arraystretch}{1.1}
\setlength{\tabcolsep}{3pt}
\begin{tabular*}{\columnwidth}{@{\extracolsep{\fill}}p{3.7cm}rr@{}}
\toprule
Criterion & Majority pass & Unanimous pass \\
\midrule
Overall usability & 126 / 126 & 122 / 126 \\
\addlinespace[2pt]
Temporal validity & 126 / 126 & 126 / 126 \\
Financial relevance & 126 / 126 & 125 / 126 \\
Dependence on new information & 126 / 126 & 126 / 126 \\
Evidence sufficiency & 126 / 126 & 125 / 126 \\
Question and option design & 126 / 126 & 124 / 126 \\
\bottomrule
\end{tabular*}
\endgroup

\caption{Human validation of final report-based questions. Overall usability requires a question to pass all five criteria.}
\label{tab:human_alignment_breakdown}
\end{table}

\FloatBarrier
\section{Boundary Estimation}
\label{sec:appendix_boundary_methods}
\label{sec:appendix_benchmark_robustness}

This section provides the formal definition of FinCut Boundary, its localization stability, and control analyses for benchmark construction. Axis-specific results are reported in Appendix~\ref{sec:appendix_axis_analysis}.

\subsection{Estimation Procedure}

\subsubsection{Observation Windows}
\label{sec:appendix_window_input}

Let $\mathrm{rel}$ be the release month of the model version actually tested, and let $\mathrm{data}_{\mathrm{start}}$ and $\mathrm{data}_{\mathrm{end}}$ be the earliest and latest months in the question bank used for evaluation. These three dates are fixed before the procedure begins, and all month offsets are calculated in calendar months. The evaluation plan specifies the observation range and the maximum number of windows $J_{\max}$ available under expansion. At check $1\leq j\leq J_{\max}$, the closed-interval window is
\begin{equation}
\begin{aligned}
\mathrm{lo}_j&=\max\{\mathrm{data}_{\mathrm{start}},\mathrm{rel}-12-6(j-1)\},\\
\mathrm{hi}&=\min\{\mathrm{rel},\mathrm{data}_{\mathrm{end}}\},\\
\mathrm{win}_j&=[\mathrm{lo}_j,\mathrm{hi}].
\end{aligned}
\label{eq:boundary_windows}
\end{equation}
Here, $\mathrm{lo}_j$ and $\mathrm{hi}$ are the first and last months of the window. The default window is $\mathrm{win}_1$. When question coverage is complete, the window lengths are 13, 19, 25 months, and so on until the procedure stops. If the next expansion would extend beyond $\mathrm{data}_{\mathrm{start}}$, we include all remaining months and record the actual number of months added. $\mathrm{data}_{\mathrm{start}},\mathrm{data}_{\mathrm{end}}$ are determined by question coverage; missing model responses do not change these limits. A window is marked as provisional when $\mathrm{data}_{\mathrm{end}}<\mathrm{rel}$. Release dates for the tested model versions are listed in Appendix~\ref{sec:appendix_experimental_setup}.

Each month contains 200 questions overall and 50 per axis. The score for question $i$ is $\mathrm{score}_i=\frac{1}{3}\sum_{q=1}^{3}\mathbf{1}\{\widehat{\mathrm{ans}}_{iq}=\mathrm{ans}_i\}$, where $\mathrm{ans}_i$ is the correct answer and $\widehat{\mathrm{ans}}_{iq}$ is the response in round $q$. Questions receive equal weight both overall and within each axis. The three responses enter statistical inference as one question-level score. Computation uses only complete windows containing all required months, questions, and valid response rounds.

Missing responses within the declared range retain their missing-data status. Months outside that range are not pending inputs for the current analysis.

\subsubsection{Decline Test}
\label{sec:appendix_decline_test}

Within window $j$, the measurement scope $s$ is either the overall question set or one of the four information axes. Each split $k$ requires at least two months on each side. We omit the window index $j$ below and use $\mu_{\mathrm{early},s,k},\mu_{\mathrm{late},s,k}$ to denote the expected mean question scores in the early and late segments. At each split, we test
\begin{equation}
\begin{aligned}
H_{0,s,k}:\;&\mu_{\mathrm{early},s,k}\leq0.25\ \text{or}\ \mu_{\mathrm{early},s,k}\leq\mu_{\mathrm{late},s,k},\\
H_{1,s,k}:\;&\mu_{\mathrm{early},s,k}>0.25\ \text{and}\ \mu_{\mathrm{early},s,k}>\mu_{\mathrm{late},s,k}.
\end{aligned}
\label{eq:boundary_hypotheses}
\end{equation}
Questions are grouped by their recorded fact identifiers. Arbitrary dependence is allowed within each fact group, and inference assumes independence between fact groups. For an observed weighted sum $\mathrm{stat}$ and the null hypothesis $\mathbb{E}[\mathrm{stat}]\leq \mathrm{null}$, let $\mathrm{rngsq}$ be the sum of squared range widths of the contributions from independent fact groups. We use Hoeffding's inequality for bounded variables to construct a conservative one-sided $p$-value:
\begin{equation}
p(\mathrm{stat};\mathrm{null})=\exp\!\left[-\frac{2\max(\mathrm{stat}-\mathrm{null},0)^2}{\mathrm{rngsq}}\right].
\end{equation}
Let $N_{\mathrm{early}},N_{\mathrm{late}}$ be the numbers of questions in the two segments, and let $n_{\mathrm{early},g},n_{\mathrm{late},g}$ be the corresponding numbers for fact group $g$. The early mean and the early-minus-late difference use, respectively,
\begin{equation}
\begin{aligned}
\mathrm{rngsq}_{\mathrm{early}}&=\sum_g\left(\frac{n_{\mathrm{early},g}}{N_{\mathrm{early}}}\right)^2,\\
\mathrm{rngsq}_{\mathrm{diff}}&=\sum_g\left(\frac{n_{\mathrm{early},g}}{N_{\mathrm{early}}}+\frac{n_{\mathrm{late},g}}{N_{\mathrm{late}}}\right)^2.
\end{aligned}
\end{equation}
The early sample mean uses $\mathrm{null}=0.25$ and $\mathrm{rngsq}_{\mathrm{early}}$, while the difference between the early and late sample means uses $\mathrm{null}=0$ and $\mathrm{rngsq}_{\mathrm{diff}}$. Both one-sided conditions must hold, so the combined $p$-value for each split is the larger of the two. This step does not require the two statistics to be independent. The overall question set and each information axis form separate testing families. For a fixed scope $s$, we apply Holm correction to the $p$-values of all splits in the current window. Check $j$ uses
\begin{equation}
\alpha_j=\frac{0.05}{j(j+1)},\qquad\sum_{j=1}^{\infty}\alpha_j=0.05.
\end{equation}
A fixed scope $s$ passes the decline test if at least one split has a corrected $p\leq\alpha_j$. The initial window uses $\alpha_1=0.025$. Combining Holm correction within each window with the significance budget across windows keeps the false positive probability of the overall decline test at or below 0.05 throughout the search for each model. This result follows from the union bound and therefore also applies to dependent windows. Each axis uses the same correction rule and the budget assigned to the selected window; its false positive control applies separately to that axis. The overall result determines the stopping window, and axis diagnostics are reported within that window. The decline test and knot consistency checks under the two minimum low-phase lengths determine the overall Boundary point estimate. Resampling separately describes localization stability.

\subsubsection{Boundary Fitting}

Let $n_{\mathrm{mon}}$ be the number of months in the window, indexed in chronological order as $t=1,\ldots,n_{\mathrm{mon}}$. For every scope that passes the decline test, we fit the same piecewise declining curve:
\begin{equation}
\begin{aligned}
\widehat{\mathrm{acc}}(t)={}&\mathrm{acc}_{\mathrm{lo}}+(\mathrm{acc}_{\mathrm{hi}}-\mathrm{acc}_{\mathrm{lo}})\\
&\quad\cdot\left[\frac{\mathrm{Boundary}-t}{\mathrm{Boundary}-\mathrm{Start}}\right]_0^1,
\end{aligned}
\label{eq:fixed_c_endpoint}
\end{equation}
where $\widehat{\mathrm{acc}}(t)$ is the fitted accuracy for month $t$, $[x]_0^1=\min(1,\max(0,x))$, and $0\leq \mathrm{acc}_{\mathrm{lo}}\leq \mathrm{acc}_{\mathrm{hi}}\leq1$. $\mathrm{Start}$ is the last month of the high phase, and $\mathrm{Boundary}$ is the first month in which the curve reaches the low level. The high and low levels, $\mathrm{acc}_{\mathrm{hi}},\mathrm{acc}_{\mathrm{lo}}$, are jointly estimated from the observations. The high and low phases each contain at least two months, including Start and Boundary, respectively. The transition spans at least one interval between months. We enumerate all valid Start and Boundary months and solve a constrained least-squares problem for each pair, minimizing the sum of squared errors with equal weight for every month. For errors tied within $10^{-12}$, we select the later Boundary and then the later Start.

Candidate Boundaries cannot be later than the release month. We fit the curve once with a minimum low-phase length of two months and once with three months, recording Start and Boundary from both fits. The two-month setting gives the main fit. When Start is the second observed month, the high phase is exactly at its two-month minimum and is marked as touching the left limit. A Boundary at the latest month allowed by the observation end and the minimum low-phase length is marked as touching the right limit.

\subsubsection{FinCut Boundary}
\label{sec:appendix_boundary_intervals}

After the overall question set passes the decline test and knot checks, FinCut Boundary is the first month reaching the fitted low level under the two-month minimum low-phase setting. The overall point estimate is always reported together with Localization Rate and Boundary Range. Each axis independently undergoes the decline test and knot checks within the overall stopping window. Leading-Axis Boundary is selected from axes whose Boundary Range spans at most six months. The low level is estimated from the data and may be above or below 25\%. Multiple-testing correction controls the error rate of the decline test, while full-procedure resampling assesses the stability of date localization.

Pointwise intervals for monthly accuracy and localization stability for overall and axis-specific Boundaries concern different statistical quantities. Table~\ref{tab:resampling_design} summarizes their sampling designs.

\subsubsection{Window Selection}
\label{sec:appendix_window_stopping}

For each complete window, we run the decline test and then the knot checks. After the overall decline test passes, the Boundary point estimate is determined in the current window and expansion stops when the two- and three-month minimum low-phase settings produce Start estimates that differ by at most one month and Boundary estimates that differ by at most one month, and when neither Start touches the left limit. We retain a flag if either low-phase setting touches the right limit. The knots under the two minimum low-phase lengths must still meet the same consistency requirements.

\texttt{SelectWindow} checks windows within the declared range in the order given by Eq.~\ref{eq:boundary_windows}. It returns the first window that passes the overall decline test and knot checks, together with its Boundary result. If a window fails and $J_{\max}$ has not been reached, it extends the window six months backward. When the range is exhausted, a result with no detected decline is marked ND, and a detected decline with unconfirmed knots is marked Unresolved. If a later window within the range lacks complete responses, the last complete window is retained and the result is marked Pending. The initial window must be complete before Algorithm~\ref{alg:boundary_window_expansion} is run.

\begin{algorithm}[!htbp]
\caption{Boundary estimation procedure.}
\label{alg:boundary_window_expansion}
\footnotesize
\begin{algorithmic}[1]
\Require Verified \texttt{responses} covering the initial window
\Statex \hspace{\algorithmicindent} Windows $1{:}J_{\max}$ from Eq.~\ref{eq:boundary_windows}; 1,000 bootstrap samples
\Ensure Overall and axis Boundaries, localization summaries, leading axes
\State $(\texttt{window},\texttt{overall})\gets\Call{SelectWindow}{\texttt{responses}}$
\State $\texttt{samples}\gets\Call{StratifiedResample}{\texttt{responses},1000}$
\State $\texttt{overallCopies}\gets[\ ]$
\For{each \texttt{sample} in \texttt{samples}}
    \State $\texttt{overallCopy}\gets\Call{SelectWindow}{\texttt{sample}}$
    \State append \texttt{overallCopy} to \texttt{overallCopies}
\EndFor
\State $\texttt{overallStable}\gets\Call{Summarize}{\texttt{overallCopies}}$
\State $\texttt{axisFits}\gets[\ ]$
\For{each \texttt{axis} in the four financial axes}
    \State $\texttt{axisInput}\gets\texttt{responses}|_{\texttt{window}}$
    \State $\texttt{axisFit}\gets\Call{TestAndFit}{\texttt{axisInput},\texttt{axis}}$
    \State $\texttt{axisCopies}\gets[\ ]$
    \For{each \texttt{sample} in \texttt{samples}}
        \State $\texttt{axisInput}\gets\texttt{sample}|_{\texttt{window}}$
        \State $\texttt{axisCopy}\gets\Call{TestAndFit}{\texttt{axisInput},\texttt{axis}}$
        \State append \texttt{axisCopy} to \texttt{axisCopies}
    \EndFor
    \State $\texttt{axisStable}\gets\Call{Summarize}{\texttt{axisCopies}}$
    \State append $(\texttt{axisFit},\texttt{axisStable})$ to \texttt{axisFits}
\EndFor
\State $\texttt{leadingAxes}\gets\Call{LatestTight}{\texttt{axisFits}}$
\State \Return all results and stability summaries
\end{algorithmic}
\end{algorithm}

\texttt{TestAndFit} runs the decline test, fits the curve under both minimum low-phase lengths, and checks knot consistency. \texttt{Summarize} calculates Localization Rate, Boundary Range, and the Tight/Wide label as defined in Section~\ref{sec:appendix_fbc}. The four axes are independently estimated within the fixed overall stopping window. \texttt{LatestTight} selects the latest date among Tight axes and retains ties. ND, Unresolved, and Pending retain the statuses of no detected decline, unresolved knots, and missing inputs, respectively. Sampling settings are given in Table~\ref{tab:resampling_design}.

\subsection{Stability of Boundary Estimates}

\subsubsection{Localization Stability}
\label{sec:appendix_fbc}

\textbf{Sampling units and weights.} Table~\ref{tab:resampling_design} summarizes the resampling settings. In the current question sets, each recorded fact group contains one question. The three responses to each question are averaged first and always kept together. Within each stratum, we independently draw groups with equal probability and replacement, using the same number of draws as the original number of groups. Repeatedly drawn groups contribute to the mean according to their multiplicity, while groups that are not drawn are absent from that replicate. Month order is fixed. Boundary analysis keeps 50 draws for each of the four axes in each month; language and question source are not used as additional strata.

\begin{table*}[!tp]
\centering\small
\setlength{\tabcolsep}{3pt}
\renewcommand{\arraystretch}{1.15}
\begin{tabular}{@{}p{.16\textwidth}p{.38\textwidth}p{.12\textwidth}p{.27\textwidth}@{}}
\toprule
Quantity & Resampling design & Replicates & Output \\
\midrule
Monthly accuracy & Within each month: 200 draws from all 200 groups, or 50 from the selected axis's 50 groups & Exact distribution & Pointwise 95\% percentile interval \\
Overall Boundary & Within each month-by-axis stratum: 50 draws from 50 groups; shared across nested windows & 1,000 & Localization Rate and Boundary Range \\
Axis Boundary & Same draws as above; each axis evaluated in the full-data overall stopping window & Same 1,000 & Axis Localization Rate and Boundary Range \\
Matched difficulty control & Within each month: $n$ draws from $n$ groups in the analyzed subset; shared across target and control models & 5,000 & Paired early-minus-late contrasts and pointwise 95\% intervals \\
\bottomrule
\end{tabular}
\caption{Resampling designs. All draws are uniform with replacement and preserve the original stratum size.}
\label{tab:resampling_design}
\end{table*}

\textbf{Monthly intervals and paired controls.} Monthly intervals in the main experiment use the empirical distribution of three-response question means $\{0,1/3,2/3,1\}$. Discrete convolution gives the exact distribution of the mean under sampling with replacement at the original sample size, from which we take the 2.5\% and 97.5\% quantiles. Overall intervals pool all 200 questions in a month, whereas Boundary resampling fixes the sample size for each axis. The difficulty control uses shared sampling indices for target and control models within each month, preserving question-level pairing and each model's original three-response mean or single-response score. We first calculate monthly means and then take the difference between the equally weighted averages of the first and last four months. The 2.5\% and 97.5\% quantiles of its 5,000 replicates use linear interpolation and form exploratory pointwise intervals.

\textbf{Procedure replay and denominators.} Each replicate in the overall analysis reruns the decline test, fitting and knot consistency checks under both minimum low-phase lengths, and window expansion and stopping. Nested windows share the question scores from that replicate. Axis analysis fixes the overall window selected from the complete data and reruns each axis's tests and fitting. Each replicate uses a single level of resampling. Localization Rate, the exact recurrence rate, and the recurrence rate within one month all use the full set of 1,000 replicates as the denominator. Boundary Range uses only replicates that yield a Boundary.

\textbf{Completeness and reproducibility.} Computation uses verified complete months. The main experiment requires 200 questions per month, 50 per axis, and valid responses in all three rounds. Paired controls require responses from all selected models to the same questions; missing inputs are not replaced by smaller samples. An overall replicate is marked Pending when it needs to expand to months within the declared range that lack complete responses. A replicate that reaches the lower limit of the range without passing is marked ND or Unresolved. All such replicates remain in the denominator. Random sampling uses NumPy's default generator with a base seed of 2048. Seeds are deterministically derived from model or comparison identifiers; the exact rules are provided in the reproduction code.

\textbf{Localization Rate} is the proportion of the 1,000 replicates that pass the decline test and knot checks and yield a Boundary. \textbf{Boundary Range} is the empirical 5\%--95\% quantile range of Boundary dates among these localized replicates. This range is conditional on successful localization and therefore needs to be interpreted together with Localization Rate. A span of at most six months is labeled Tight, and a span of more than six months is labeled Wide. Overall results report the point estimate, Localization Rate, and Boundary Range together. For axis diagnostics, Tight axes are eligible for Leading-Axis Boundary; we select the latest date among them and retain ties.

Table~\ref{tab:transition_pair_stability} reports complete results for the twelve models in the main table and supplementary analysis. Among the nine models in the main table with a Boundary, Localization Rate ranges from 89.6\% to 100\%. Qwen3.5-4B shows no detected decline in the complete data, and only 7.2\% of its 1,000 replicates yield a Boundary. After adding observations from October 2023 to March 2024 for Haiku 4.5, its Localization Rate is 89.6\%, and its Boundary Range is July 2024 to April 2025. Qwen3.5-122B has a Localization Rate of 99.6\% and a Boundary Range of February 2025 to January 2026. Localization Rate shows whether a Boundary can be located again, while Boundary Range shows how concentrated its date is among successful replicates.

\begin{table*}[!tp]
\centering\scriptsize
\begingroup
\renewcommand{\arraystretch}{1.08}
\setlength{\tabcolsep}{3pt}
\begin{tabular}{@{}lrrrrll@{}}
\toprule
Model & Trials & Localization Rate & Exact Boundary & Within $\pm1$ month & Boundary Range & Width \\
\midrule
DeepSeek-V4.1-Flash-1017 & 1000 & 96.2\% & 58.2\% & 95.6\% & 2026-06--2026-07 & Tight \\
Qwen3.8-27B & 1000 & 100.0\% & 72.0\% & 87.0\% & 2026-02--2026-06 & Tight \\
Qwen3.5-4B & 1000 & 7.2\% & -- & -- & -- & -- \\
Qwen3.5-122B-A10B & 1000 & 99.6\% & 15.3\% & 38.2\% & 2025-02--2026-01 & Wide \\
Qwen3.5-397B-A17B & 1000 & 100.0\% & 23.9\% & 50.1\% & 2025-02--2025-07 & Tight \\
GPT-6-Astra & 1000 & 100.0\% & 97.4\% & 100.0\% & 2026-06--2026-06 & Tight \\
GPT-5.6-Sol & 1000 & 90.2\% & 32.9\% & 38.5\% & 2026-03--2026-06 & Tight \\
Gemini-3.8-Flash & 1000 & 99.9\% & 68.7\% & 75.5\% & 2026-03--2026-06 & Tight \\
Claude-Haiku-5.5 & 1000 & 99.6\% & 64.9\% & 81.5\% & 2026-03--2026-07 & Tight \\
Claude-Haiku-4.5 & 1000 & 89.6\% & 45.7\% & 74.3\% & 2024-07--2025-04 & Wide \\
Qwen3.5-27B & 1000 & 99.7\% & 28.6\% & 35.9\% & 2025-02--2026-01 & Wide \\
GPT-5.6-Luna & 1000 & 98.8\% & 92.9\% & 96.1\% & 2026-05--2026-06 & Tight \\
\bottomrule
\end{tabular}
\endgroup

\caption{Full-procedure resampling of window selection. Localization Rate, Exact Boundary, and Within $\pm1$ month use all 1,000 replicates as the denominator. Boundary Range is the middle 90\% of dates among replicates yielding a Boundary. Tight spans at most six months; Wide spans more.}
\label{tab:transition_pair_stability}
\end{table*}

\subsubsection{Qwen3.5 Main Models}
\label{sec:appendix_qwen35_profiles}

Some Qwen3.5 models show a long, gradual decline with month-to-month fluctuations. For 4B, no decline is detected in the complete data from January 2023 to March 2026, and the resampling Localization Rate is 7.2\%; we therefore do not report a Boundary Range. For 122B, a decline is detected, with a FinCut Boundary point estimate of December 2025, a Localization Rate of 99.6\%, and a Boundary Range from February 2025 to January 2026, yielding a Wide label. For 397B, the point estimate is April 2025 and the Boundary Range is February to July 2025, yielding a Tight label. These results distinguish evidence of decline, the point estimate from complete data, and the precision of date localization.

\subsection{Benchmark Construction Checks}
\label{sec:appendix_construction_robustness}

This section uses existing responses to the same question sets to examine how cross-month differences in question difficulty and question generation methods relate to accuracy decline. Questions, answers, and boundary estimation rules remain the same.

\subsubsection{Question Difficulty across Months}
\label{sec:appendix_difficulty_drift}

Each of the 45 monthly question sets contains 200 questions: 70 from research reports, 30 from indicators, and 100 from market data. Each language--axis combination contains 25 questions, and this composition is held fixed across months. We further use fixed versions of Qwen2.5~\cite{qwen25} at 7B, 14B, 32B, and 72B as controls, asking them the same questions from the same months. These versions were released before the observation windows covered by this control analysis. The controls use existing single-response scores, while the main models use three-response means for each question.

Table~\ref{tab:difficulty_control_contrasts} compares mean accuracy in the first and last four months of each default window. Difference is the main model's decline minus the control decline; paired sampling and interval settings follow Table~\ref{tab:resampling_design}. All five main models decline more than the controls, and all pointwise 95\% confidence intervals for the differences are above zero.

\begin{table}[!htbp]
\centering\small
\begingroup
\renewcommand{\arraystretch}{1.08}
\setlength{\tabcolsep}{2pt}
\begin{tabular*}{\columnwidth}{@{\extracolsep{\fill}}lrrr@{}}
\toprule
Model & \shortstack[r]{Target\\drop} & \shortstack[r]{Control\\drop} & \shortstack[r]{Difference\\{[95\% CI]}} \\
\midrule
GPT-5.6-Sol & 46.8 & 3.3 & 43.5 [39.2, 47.7] \\
GPT-6-Astra & 50.2 & 3.9 & 46.3 [42.2, 50.5] \\
Gemini-3.8-Flash & 10.1 & 3.9 & 6.2 [2.0, 10.5] \\
DeepSeek-V4.1-Flash-1017 & 34.8 & 4.5 & 30.3 [26.1, 34.4] \\
Qwen3.8-27B & 13.9 & 3.7 & 10.2 [6.3, 14.0] \\
\bottomrule
\end{tabular*}
\endgroup

\caption{Early-minus-late accuracy changes in percentage points, using the first and last four months of each default window. Difference subtracts the mean control change from the target change. Exploratory pointwise 95\% intervals follow Table~\ref{tab:resampling_design}.}
\label{tab:difficulty_control_contrasts}
\end{table}

\subsubsection{Question-Generator Preferences}
\label{sec:appendix_generator_preferences}

Each month, 130 time-series questions are generated programmatically, and 70 report-based questions are generated by GPT and checked in independent contexts. The human review protocol is given in Appendix~\ref{sec:appendix_human_alignment_protocol}. When results are split by source, Sol, Astra, and DeepSeek all show declines on report, indicator, and market questions, showing that this temporal change occurs across different generation methods.

Over the shared period from January to September 2026, Sol's advantage over Gemini and Qwen3.8 appears on both programmatically generated and report-based questions. Non-GPT families also achieve high accuracy on recent report-based questions.

\FloatBarrier
\section{Experimental Setup}
\label{sec:appendix_experimental_setup}

This section provides model versions, inference settings, and the prompts used in evaluation. Statistical methods are described in Appendix~\ref{sec:appendix_boundary_methods}.

\subsection{Models}

Qwen3.5 and Qwen3.8 run through local vLLM services. DeepSeek-V4.1-Flash-1017 is accessed through its official API, while Gemini-3.8-Flash, Claude-Haiku-4.5, and Claude-Haiku-5.5 are accessed through OpenRouter, with Anthropic fixed as the provider for Claude. GPT models are evaluated through the Codex runtime.

GPT knowledge measurement uses Codex CLI 0.163.0 authenticated through a ChatGPT login. The requested model identifiers in the main table are \texttt{gpt-5.6-sol} and \texttt{gpt-6-astra}. According to the run records, Sol's Closed-Choice and IDK experiments were conducted on October 22--23, 2026, and Astra's on October 22--25. The supplementary GPT-5.6-Luna experiments were conducted on October 22--23. All dates are in UTC. The access route and experiment dates jointly specify the tested service version; a model under the same name may be updated later.

The Claude experiments use \texttt{anthropic/claude-haiku-4.5} and \texttt{anthropic/claude-haiku-5.5} on OpenRouter. Their recorded provider versions are \texttt{anthropic/claude-4.5-haiku-20251001} and \texttt{anthropic/claude-haiku-5.5-20261007}, respectively. Their public release dates are October 15, 2025, and October 7, 2026.

Qwen3.5-397B-A17B in the main table uses FP8 weights. Qwen3.5-4B, 122B-A10B, and Qwen3.8-27B use BF16 weights. For remotely accessed models, the tested version is specified by the model identifier and the actual access dates.

Release dates follow the release records for the tested versions. For GPT-5.6, we use the July release record corresponding to Codex. Macro, Sector, Path, and Asset denote macroeconomics, policy, and public affairs; sectors and supply chains; market paths; and individual asset moves, respectively.

\begin{table*}[!tp]
\centering\small
\begingroup
\renewcommand{\arraystretch}{1.08}
\setlength{\tabcolsep}{3.8pt}
\begin{tabular*}{\textwidth}{@{\extracolsep{\fill}}p{.38\textwidth}p{.22\textwidth}p{.34\textwidth}@{}}
\toprule
Model & Release Date & Input Window \\
\midrule
DeepSeek-V4.1-Flash-1017 & 2026-10-17 & 2025-04--2026-09 \\
Qwen3.8-27B & 2026-08-14 & 2025-08--2026-08 \\
Qwen3.5-4B & 2026-03-02 & 2023-01--2026-03 \\
Qwen3.5-122B-A10B & 2026-02-24 & 2024-02--2026-02 \\
Qwen3.5-397B-A17B & 2026-02-16 & 2024-08--2026-02 \\
GPT-6-Astra & 2026-09-03 & 2025-09--2026-09 \\
GPT-5.6-Sol & 2026-07-09 & 2025-01--2026-07 \\
Gemini-3.8-Flash & 2026-09-02 & 2025-09--2026-09 \\
Claude-Haiku-5.5 & 2026-10-07 & 2025-04--2026-09 \\
Claude-Haiku-4.5 & 2025-10-15 & 2024-04--2025-10 \\
\bottomrule
\end{tabular*}
\endgroup

\caption{Model release dates and observation windows used to estimate Transition Start and FinCut Boundary. Each window lists the complete months used for the reported Boundary estimate or no-decline assessment.}
\label{tab:main_results_coverage}
\end{table*}

\subsection{Inference Protocol}

Closed-Choice is the main evaluation setting, and IDK-Enabled is used for supplementary experiments on selected models and months. The default settings are temperature 0 and a maximum output length of 16K (16,384) tokens. Models that support explicit reasoning use the lowest enabled level, such as low, or on when only a toggle is available. Each question receives three independent responses under the evaluated setting. We first average the three correctness indicators for each question and then average across questions with equal weight.

Sampling penalties are disabled by default. For a self-hosted model, if more than 5\% of questions for any combination of month, setting, and round produce invalid first outputs with long repetition loops, we set its repetition penalty to 1.1 for all runs. We then rerun all included settings and all three rounds from the earliest month, and use only the rerun branch for the reported results. DeepSeek's thinking mode ignores the temperature parameter. In our Codex setup, temperature, output limit, and random seed are controlled by the server. Claude-Haiku-4.5 uses a fixed thinking budget of 1,024 tokens and top-p=1, while Claude-Haiku-5.5 uses adaptive reasoning at the low level. Both have a total output limit of 16,384 tokens. Temperature, random seed, and sampling penalties are omitted where the access route does not support them.

Qwen3.5-4B, 122B-A10B, and 397B-A17B in the main table all triggered the repetition rule above, so their reported results all use a repetition penalty of 1.1. Qwen3.8-27B uses the default value of 1.0. All four models use temperature 0 and top-p=1, with the top-k limit and all other sampling penalties disabled. The request seeds for the three rounds are 2048, 2049, and 2050. Qwen3.5 runs with thinking enabled, and Qwen3.8 uses the low level. These results correspond to the recorded deployment precision and inference settings for each model.

Each question uses an independent context. The task input contains the question, four options, and the response instructions for the relevant setting. It excludes evidence, explanations, correct answers, and conversations from other questions. GPT experiments call Codex independently for each question. The parser accepts only the specified answer set. Invalid or truncated outputs are retried up to 20 times by default and up to 3 times in the Codex experiments. Questions with outputs that remain invalid are kept for manual review.

Closed-Choice requires the model to select A, B, C, or D. IDK-Enabled additionally permits IDK. The following prompts are exported verbatim from the actual evaluation templates. The two placeholders are replaced by the question and the four options, respectively.

\tcbinputlisting{
  enhanced,breakable,
  colback=white,colframe=Gray!50!black,
  colbacktitle=Gray!60!black,coltitle=white,
  title=Closed-Choice Prompt,
  listing engine=listings,listing only,
  listing file=prompts/closed_choice.txt,
  listing options={basicstyle=\ttfamily\scriptsize,
    breaklines=true,columns=fullflexible,keepspaces=true,
    showstringspaces=false}
}

\tcbinputlisting{
  enhanced,breakable,
  colback=white,colframe=Gray!50!black,
  colbacktitle=Gray!60!black,coltitle=white,
  title=IDK-Enabled Prompt,
  listing engine=listings,listing only,
  listing file=prompts/idk_enabled.txt,
  listing options={basicstyle=\ttfamily\scriptsize,
    breaklines=true,columns=fullflexible,keepspaces=true,
    showstringspaces=false}
}

\subsection{Compute Environment}

Local model experiments run on Ubuntu servers with four or eight NVIDIA A800 GPUs, using vLLM for inference.

\FloatBarrier
\section{Supplementary Experimental Results}
\label{sec:appendix_supplementary_results}

This section reports comparisons with vendor disclosures, results by information axis and market, and performance when abstention is allowed.

\subsection{Vendor Disclosures}
\label{sec:appendix_vendor_cutoffs}

Table~\ref{tab:official_cutoff_comparison} covers all ten models in the main table. It distinguishes general knowledge cutoffs, explicitly labeled reliable knowledge cutoffs, and training-data cutoffs, and compares them with the overall FinCut Boundary. Dates retain the precision of the official disclosures. Missing entries indicate that no date applicable to the model was found in the official sources examined. Model release dates are listed separately in the main table.

\begin{table}[!htbp]
\centering\small
\begingroup
\renewcommand{\arraystretch}{1.08}
\setlength{\tabcolsep}{4pt}
\begin{tabular}{@{}lccccc@{}}
\toprule
Model & \shortstack{Knowledge\\cutoff} & \shortstack{Reliable knowledge\\cutoff} & \shortstack{Training data\\cutoff} & \shortstack{FinCut\\Boundary} & $\Delta$ (mo.) \\
\midrule
DeepSeek-V4.1-Flash-1017 & -- & -- & Partial & 2026-06 & -- \\
Qwen3.8-27B & -- & -- & -- & 2026-03 & -- \\
Qwen3.5-4B & -- & -- & -- & ND & -- \\
Qwen3.5-122B-A10B & -- & -- & -- & 2025-12 & -- \\
Qwen3.5-397B-A17B & -- & -- & -- & 2025-04 & -- \\
GPT-6-Astra & 2026-04-30 & -- & Partial & 2026-06 & +2 \\
GPT-5.6-Sol & 2026-02-16 & -- & -- & 2026-06 & +4 \\
Gemini-3.8-Flash & 2026-03\textsuperscript{c} & -- & -- & 2026-03 & 0 \\
Claude-Haiku-5.5 & 2026-06 & 2026-06 & 2026-06 & 2026-06 & 0 \\
Claude-Haiku-4.5 & 2025-02 & 2025-02 & 2025-07 & 2025-02 & 0 \\
\bottomrule
\end{tabular}
\endgroup

\caption{Vendor disclosures and measured FinCut Boundaries. $\Delta$ is Boundary minus the reported knowledge-cutoff month; -- denotes no applicable date found, and Partial denotes a limited-scope training disclosure. \textsuperscript{c} retains Gemini's January 2025 limitation for some domains.}
\label{tab:official_cutoff_comparison}
\end{table}

\textbf{Claude's and Gemini's official disclosures are closely aligned with their measured knowledge-cutoff months.} Claude explicitly distinguishes reliable knowledge from training materials. For Haiku 4.5, the reliable-knowledge date is February 2025, the training-data date is July 2025, and the measured Boundary is February. For Haiku 5.5, both disclosed dates and the measured Boundary are June 2026~\cite{fincut_claude_haiku45_api,fincut_claude_haiku55_api}. Gemini's measured Boundary and general knowledge cutoff are both March 2026. Its official disclosure also states that some domains are covered only through January 2025~\cite{googledeepmind2026gemini38flashcard}.

\textbf{GPT's official knowledge dates are relatively conservative compared with our measurements.} The official dates for Sol and Astra are February 16 and April 30, 2026, respectively. Both have a measured Boundary in June, which is 4 and 2 calendar months later, respectively~\cite{fincut_gpt56_sol_api,fincut_gpt6_astra_api}. The monthly curves show that their advantage in answering financial facts can extend beyond the disclosed dates. Astra also discloses that some training data were collected no later than August 6, 2026~\cite{fincut_cut_gpt6_astra_training_summary}.

\textbf{Comparable dates remain limited for open-weight models.} The Qwen sources examined do not give an explicit knowledge cutoff or an overall training-data cutoff for the four tested models~\cite{fincut_qwen35,fincut_qwen38}. The training summary for the base DeepSeek V4.1 only gives a May 2025 date for some materials and a crawl period of approximately 2024--2026; an overall cutoff for the 1017 update remains unspecified~\cite{fincut_deepseek_v41_training_summary}. These differences in disclosure scope and granularity motivate a consistent behavioral measurement.

\subsection{Financial-Axis Diagnostics}
\label{sec:appendix_axis_analysis}

Table~\ref{tab:boundary_axes} reports each axis's Boundary and Boundary Range width label using the method in Appendix~\ref{sec:appendix_boundary_methods}. These results provide references for the information categories relevant to specialized tasks. Haiku 4.5's macro/policy axis shows a detected decline, but Start touches the left limit of the window. No decline is detected on its other three axes. The sector/supply-chain axis of 397B passes the knot checks, but its Boundary Range is Wide, so it is not a Leading-Axis candidate.

\begin{table}[!ht]
\centering\footnotesize
\resizebox{\textwidth}{!}{
\begingroup
\renewcommand{\arraystretch}{1.08}
\setlength{\tabcolsep}{3.5pt}
\begin{tabular}{@{}lcrcrcrcr@{}}
\toprule
Model & Macro & Width & Sector & Width & Path & Width & Asset & Width \\
\midrule
Qwen3.5-122B-A10B & 2025-03 & Tight & Unresolved & -- & Unresolved & -- & ND & -- \\
Qwen3.5-397B-A17B & 2025-04 & Tight & 2025-08 & Wide & Unresolved & -- & ND & -- \\
Qwen3.8-27B & 2026-03 & Tight & ND & -- & ND & -- & ND & -- \\
DeepSeek-V4.1-Flash-1017 & 2026-08\textsuperscript{b} & Tight & 2026-07\textsuperscript{b} & Tight & 2026-04 & Tight & Unresolved & -- \\
GPT-5.6-Sol & 2026-06\textsuperscript{b} & Tight & 2026-03 & Tight & 2026-04 & Tight & 2026-06\textsuperscript{b} & Tight \\
GPT-6-Astra & 2026-07\textsuperscript{b} & Tight & 2026-06 & Tight & 2026-05 & Tight & 2026-06 & Tight \\
Gemini-3.8-Flash & 2026-03 & Tight & ND & -- & ND & -- & ND & -- \\
Claude-Haiku-4.5 & Unresolved & -- & ND & -- & ND & -- & ND & -- \\
Claude-Haiku-5.5 & 2026-08\textsuperscript{b} & Tight & 2026-06 & Tight & 2026-03 & Tight & ND & -- \\
\bottomrule
\end{tabular}
\endgroup
}
\caption{Axis-specific Boundaries and Boundary Range width in the fixed overall stopping window. Tight spans at most six months and Wide spans more. ND denotes no significant decline; Unresolved denotes a detected decline with unconfirmed knots. Only Tight axes are eligible for Leading-Axis Boundary. \textsuperscript{b} marks right-limit contact of either fitted Boundary.}
\label{tab:boundary_axes}
\end{table}

\subsection{Market-Specific Performance}
\label{sec:appendix_market_boundaries}

We evaluate Chinese questions about the Chinese market and English questions about the U.S. market separately. Each market has 100 questions per month, with 25 per axis and three responses per question. The two markets select their own stopping windows, and each market's four axes are independently checked within its window, following the tests, fitting, and stability summaries in Appendix~\ref{sec:appendix_boundary_methods}. Accuracy comparisons use each model's initial window, covering the 12 months before release and the release month. For models whose release month is not yet covered by the question bank, the window ends in September 2026. We first average the three responses for each question, apply a two-sided Hoeffding test to the accuracy difference between the two markets, and then apply Holm correction across nine models. Table~\ref{tab:market_boundary_comparison} summarizes accuracies, differences, and overall and leading-axis results.

\begin{table}[!ht]
\centering\footnotesize
\resizebox{\textwidth}{!}{
\begingroup
\renewcommand{\arraystretch}{1.08}
\setlength{\tabcolsep}{2.5pt}
\begin{tabular}{@{}llrcrclrr@{}}
\toprule
Model & Market & Acc. & \shortstack{Overall FinCut\\Boundary} & Width & \shortstack{Leading-Axis\\Boundary} & Axis Width & $\Delta$Acc. & $p_{\mathrm{Holm}}$ \\
\midrule
\multicolumn{9}{@{}l}{\textbf{Open-Weight Models}} \\
DeepSeek-V4.1-Flash-1017\textsuperscript{a} & CN & 44.11 & 2026-07 & Tight & 2026-06 & Macro/Path: Tight & $ +1.25 $ & $1.0000$ \\
 & US & 45.36 & 2026-06 & Tight & 2026-08 & Macro: Tight &  &  \\
\addlinespace[2pt]
Qwen3.8-27B & CN & 31.46 & 2026-03 & Tight & 2026-03 & Macro: Tight & $ -0.15 $ & $1.0000$ \\
 & US & 31.31 & 2026-02 & Tight & ND & -- &  &  \\
\addlinespace[2pt]
Qwen3.5-122B-A10B & CN & 28.10 & 2025-08 & Wide & 2025-02 & Macro: Tight & $ +1.28 $ & $1.0000$ \\
 & US & 29.38 & 2026-01 & Wide & \shortstack{Not\\identified} & -- &  &  \\
\addlinespace[2pt]
Qwen3.5-397B-A17B & CN & 28.00 & 2025-07 & Tight & 2025-03 & Macro: Tight & $ +3.18 $ & $1.0000$ \\
 & US & 31.18 & 2025-04 & Wide & 2025-05 & Macro: Tight &  &  \\
\addlinespace[2pt]
\addlinespace[3pt]
\multicolumn{9}{@{}l}{\textbf{Closed-Weight Models}} \\
GPT-6-Astra & CN & 57.36 & 2026-06 & Tight & 2026-06 & Macro/Sector/Path: Tight & $ +9.74 $ & $<0.0001^{***}$ \\
 & US & 67.10 & 2026-06 & Tight & 2026-08 & Macro: Tight &  &  \\
\addlinespace[2pt]
GPT-5.6-Sol & CN & 54.41 & 2026-06 & Tight & 2026-06 & Macro: Tight & $ +10.00 $ & $<0.0001^{***}$ \\
 & US & 64.41 & 2026-03 & Tight & 2026-04 & Macro: Tight &  &  \\
\addlinespace[2pt]
Gemini-3.8-Flash & CN & 33.59 & 2026-05 & Tight & 2026-03 & Macro: Tight & $ +2.90 $ & $1.0000$ \\
 & US & 36.49 & 2026-03 & Tight & 2026-04 & Macro: Tight &  &  \\
\addlinespace[2pt]
Claude-Haiku-5.5\textsuperscript{a} & CN & 34.56 & 2026-03 & Tight & 2026-08 & Macro: Tight & $ +7.14 $ & $0.0309^{**}$ \\
 & US & 41.69 & 2026-05 & Tight & 2026-08 & Macro: Tight &  &  \\
\addlinespace[2pt]
Claude-Haiku-4.5 & CN & 28.95 & ND & -- & ND & -- & $ +3.05 $ & $1.0000$ \\
 & US & 32.00 & ND & -- & \shortstack{Not\\identified} & -- &  &  \\
\bottomrule
\end{tabular}
\endgroup
}
\caption{Market-specific accuracy and FinCut Boundaries. Acc. is a percentage; $\Delta$Acc.$=\mathrm{US}-\mathrm{CN}$ is in percentage points. $p_{\mathrm{Holm}}$ is corrected across nine models, with * $p<0.10$, ** $p<0.05$, and *** $p<0.01$. Width is Tight for a Boundary Range of at most six months and Wide otherwise; Axis Width names each leading axis and its tag. ND denotes no detected decline; Not identified denotes that no axis qualifies for Leading-Axis Boundary. \textsuperscript{a} marks a provisional release window.}
\label{tab:market_boundary_comparison}
\end{table}

\textbf{The English-question accuracy advantage is concentrated in some models.} Astra, Sol, and Haiku 5.5 have average accuracies that are 9.7, 10.0, and 7.1 percentage points higher, respectively, on English questions about the U.S. market. All three differences reach the 5\% significance level after correction across nine models; the other models do not meet this threshold. Mean accuracy and the end of the decline provide complementary information. For example, Sol has higher English accuracy, but its English Boundary is earlier than its Chinese Boundary.

\textbf{Some models show differences consistent with coverage of their home markets.} DeepSeek and Qwen3.8 both have Chinese overall Boundaries one month later than their English Boundaries, while Haiku 5.5's English Boundary is two months later than its Chinese Boundary. These differences may relate to language and market coverage in the training data. The direction varies by model: Sol and Gemini also have later Chinese Boundaries, while Astra has the same Boundary in both markets. Each language corresponds to one market in the question sets, so these results describe differences in knowledge coverage between the two market-specific question sets.

\textbf{Market-specific estimates complement the overall results.} For both Chinese and English, 8/9 models show a detected decline and yield a FinCut Boundary point estimate. With half as many questions per market, detection power also changes. Haiku 4.5 shows no significant decline in either market within the April 2024--October 2025 observation range and is therefore marked ND. Qwen3.5-122B has point estimates of August 2025 for the Chinese market and January 2026 for the U.S. market; both Boundary Ranges are Wide. For tasks focused on a particular market, the corresponding market- and axis-specific Boundaries and width labels can guide the choice of reference periods.

\subsection{Abstention with IDK-Enabled}
\label{sec:idk_results}

We compare Closed-Choice and IDK-Enabled on the same questions from January to September 2026, with three independent responses per question in each setting. Table~\ref{tab:idk_comparison} shows that allowing abstention lowers overall accuracy for all five models while increasing accuracy on responses that select A--D. GPT-6-Astra is especially likely to answer only a small number of questions.

\begin{table}[!htbp]
\centering
\small
\begingroup
\renewcommand{\arraystretch}{1.15}
\setlength{\tabcolsep}{2.5pt}
\begin{tabular*}{\columnwidth}{@{\extracolsep{\fill}}lrrrr@{}}
\toprule
& CC & \multicolumn{3}{c}{IDK-Enabled} \\
\cmidrule(lr){3-5}
Model & Acc. & Acc. & IDK & Sel. \\
\midrule
GPT-5.6-Sol & 41.7 & 9.5 & 87.4 & 75.6 \\
GPT-5.6-Terra & 38.0 & 18.9 & 63.7 & 52.2 \\
GPT-5.6-Luna & 35.6 & 9.8 & 82.9 & 57.0 \\
GPT-6-Astra & 53.8 & 3.5 & 96.4 & 98.4 \\
DeepSeek-V4.1-Flash-1017 & 37.5 & 10.8 & 84.6 & 69.9 \\
\bottomrule
\end{tabular*}
\endgroup

\caption{Paired Closed-Choice (CC) and IDK-Enabled results on January--September 2026: 1,800 questions and three responses per question in each condition. All values are percentages. Acc. counts abstentions as incorrect; IDK is the abstention rate; Sel. is accuracy among non-abstaining responses.}
\label{tab:idk_comparison}
\end{table}

Overall accuracy under IDK counts abstentions as incorrect, while selective accuracy uses only A--D responses. The two satisfy the identity ``overall accuracy = response coverage $\times$ selective accuracy.'' Closed-Choice estimates the boundary under a common forced-choice setting, while IDK-Enabled additionally describes selective answering behavior.

\section{Downstream Experimental Details}
\label{sec:appendix_downstream}

\subsection{Observation Windows}

Let $S$ and $B$ denote the overall Start and FinCut Boundary from independent knowledge measurement, respectively. The planned months are
\begin{equation}
\begin{aligned}
W_{\mathrm{pre}}^{*}&=\{S-6,\ldots,S-1\},\\
W_{\mathrm{post}}^{*}&=\{B+1,\ldots,B+6\}.
\end{aligned}
\label{eq:downstream_windows}
\end{equation}
The actual ranking windows contain up to six consecutive complete months immediately before the Start and up to six immediately after the Boundary. Both knot months are excluded. Windows are shortened only by the first and last months covered by the question bank and the data cutoff of September 30, 2026. The two GPT models and Haiku each have three months in the later period, while Qwen3.8 has six months on both sides. Missing responses and score levels do not change the date ranges. Qwen2.5-72B uses the same actual months and questions as each target model. Table~\ref{tab:ranking_windows} lists the actual ranges for the four target models.

The quarterly earnings task uses each of the four target models' Start and Boundary to define Disclosure-Pre and Disclosure-Post from the month in which the target quarter's results were first disclosed. Qwen3.8 has a Start of December 2025 and a Boundary of March 2026, with Disclosure-Pre covering June--November 2025 and Disclosure-Post covering April--September 2026. Sol's Start is November 2025, while Luna and Haiku have a Start of October 2025; all three have a Boundary of June 2026. Sol's Disclosure-Pre window is May--October 2025, and the Disclosure-Pre windows for Luna and Haiku are April--September 2025. Earnings data extend through October 31, 2026, so all three have Disclosure-Post windows of July--October 2026. Each window retains companies with eligible questions in both periods, and Qwen2.5-72B uses the same companies, quarters, and windows. The forecasting information cutoff is the day before the target quarter begins. A common later-period comparison by forecasting decision time requires quarters that begin after each model's Boundary and have disclosed results. Each question receives three independent responses under each condition. Inference uses the lowest enabled reasoning setting, following the main experiments.

\begin{table*}[!tp]
\centering\small
\begingroup%
\scriptsize%
\setlength{\tabcolsep}{2pt}%
\renewcommand{\arraystretch}{1.12}%
\begin{tabular*}{\textwidth}{@{\extracolsep{\fill}}lllllr@{}}%
\toprule
Model & Start & Boundary & Pre span & Post span & $n_{\mathrm{pre}}/n_{\mathrm{post}}$ \\
\midrule
Qwen3.8-27B & 2025-12 & 2026-03 & 2025-06--2025-11 & 2026-04--2026-09 & 6/6 \\
GPT-5.6-Sol & 2025-11 & 2026-06 & 2025-05--2025-10 & 2026-07--2026-09 & 6/3 \\
GPT-5.6-Luna & 2025-10 & 2026-06 & 2025-04--2025-09 & 2026-07--2026-09 & 6/3 \\
Claude-Haiku-5.5 & 2025-10 & 2026-06 & 2025-04--2025-09 & 2026-07--2026-09 & 6/3 \\
\bottomrule
\end{tabular*}%
\endgroup%

\caption{Start, FinCut Boundary, and observed ranking windows. Each side contains up to six complete months; knot months are excluded. Returns are available through September 2026. Qwen2.5-72B answers identical questions in each target's windows.}
\label{tab:ranking_windows}
\end{table*}

\subsection{Asset Ranking}

China and the United States each use a fixed pool of 30 assets, comprising 15 stocks and 15 exchange-traded funds (ETFs). The assets are selected by stratified sampling with a fixed seed from an original 50-asset pool and remain the same across months and models. Questions for China are in Chinese and those for the United States are in English; each request covers only one market. The model receives Markdown tables containing integer IDs, actual tickers, names, asset types, and seven historical features: returns over the past 1, 3, and 6 months; volatility over the past 1 and 3 months; maximum drawdown over the past 3 months; and the change in trading volume over the most recent 21 trading days relative to the preceding 21 days. All seven features are converted to average-rank percentiles within the month's 30-asset pool, expressed from 0 to 100 with two decimal places. Drawdowns closer to zero receive higher percentiles. The question also provides 5-, 21-, and 63-day returns and 21-day annualized volatility for three representative indices: the CSI 300, CSI 500, and ChiNext Index for China, and the S\&P 500, Nasdaq-100, and Russell 2000 for the United States.

All features use data through the close of the last trading day before the target month $t$. The primary metric assumes that the signal is implemented at the close of the first trading day of month $t$. The target return is $\mathrm{ret}_{i,t}=\mathrm{px}^{\mathrm{adj}}_{i,\mathrm{out}}/\mathrm{px}^{\mathrm{adj}}_{i,\mathrm{in}}-1$, where in and out denote the first and last trading days of month $t$, respectively, and prices are adjusted closing prices from the same fixed data version. The model returns a JSON \texttt{ranking} containing every ID from 1 to 30 exactly once. The first, middle, and last 10 entries form the high, medium, and low groups, respectively, and the order within each group is not scored. The \texttt{rationale} field must contain a nonempty brief explanation.

Each question in each round allows an initial response followed by at most 20 format retries, with the earliest valid response used for scoring. Truncation, missing entries, out-of-range IDs, and incorrect data types trigger retries; errors in economic judgment do not. For responses that finish normally and contain only valid IDs, we collapse consecutive duplicates. The first 30 entries after this step must include all IDs exactly once, after which repeated trailing entries may be removed. Raw responses and processing records are retained. The procedure does not fill in missing entries or reorder the ranking.

In round $q$, the high, medium, and low groups receive scores $g_{i,ctq}=2,1,0$, respectively. For market $c\in\{\mathrm{CN},\mathrm{US}\}$, we compute
\begin{equation}
\begin{aligned}
\mathrm{IC}_{ctq}&=\operatorname{Spearman}(g_{i,ctq},\mathrm{ret}_{i,ct}),\\
\mathrm{IC}_{ct}&=\frac{1}{3}\sum_{q=1}^{3}\mathrm{IC}_{ctq}.
\end{aligned}
\label{eq:downstream_rank_score}
\end{equation}
Ties receive average ranks. Within each month, we first average the three rounds for each model, then compute the target model's difference from 72B as $r_{mct}=\mathrm{IC}_{mct}-\mathrm{IC}_{72\mathrm{B},ct}$. Results for China and the United States are reported separately. The pooled result (Pooled) is the equal-weight mean of the relative ICs for the two markets in the same month and requires all three rounds for both the target and control in both markets. The statistical unit is the month; neither the three response rounds nor the 30 assets increase the temporal sample size.

\subsection{Statistical Inference}

\begin{table}[!t]
\centering\scriptsize
\begingroup%
\scriptsize%
\setlength{\tabcolsep}{2pt}%
\renewcommand{\arraystretch}{1.12}%
\begin{tabular*}{\textwidth}{@{\extracolsep{\fill}}lrrrrr@{}}%
\toprule
\multicolumn{6}{l}{\textbf{(A) Market-level inference}} \\
Market & $M$ & Pre & Post & \shortstack{$\Delta$\\{[95\% CI]}} & $p_{\downarrow}$ \\
\midrule
China & 14 & 0.054 & 0.003 & \shortstack[r]{$-0.051$\\$[-0.219, 0.086]$} & $0.2482$ \\
US & 14 & 0.047 & -0.284 & \shortstack[r]{$-0.331$\\$[-0.528, -0.061]$} & $0.0049^{***}$ \\
\bottomrule
\end{tabular*}%
\endgroup%

\par\vspace{4pt}
\begingroup%
\scriptsize%
\setlength{\tabcolsep}{2pt}%
\renewcommand{\arraystretch}{1.12}%
\begin{tabular*}{\textwidth}{@{\extracolsep{\fill}}llrrrrr@{}}%
\toprule
\multicolumn{7}{l}{\textbf{(B) Model-level inference}} \\
Model & Market & Pre & Post & \shortstack{$\Delta$\\{[95\% CI]}} & $p_{\downarrow}$ & \shortstack{$n_{\mathrm{pre}}/$\\$n_{\mathrm{post}}$} \\
\midrule
Qwen3.8-27B & CN & 0.106 & -0.038 & \shortstack[r]{$-0.144$\\$[-0.328, 0.041]$} & $0.0568^{*}$ & 6/6 \\
 & US & 0.054 & -0.261 & \shortstack[r]{$-0.315$\\$[-0.539, -0.091]$} & $0.0060^{***}$ & 6/6 \\
 & Pooled & 0.080 & -0.149 & \shortstack[r]{$-0.229$\\$[-0.405, -0.054]$} & $0.0080^{***}$ & 6/6 \\
\addlinespace
GPT-5.6-Sol & CN & 0.147 & -0.031 & \shortstack[r]{$-0.179$\\$[-0.499, 0.142]$} & $0.1124$ & 6/3 \\
 & US & 0.129 & -0.326 & \shortstack[r]{$-0.456$\\$[-0.854, -0.058]$} & $0.0181^{**}$ & 6/3 \\
 & Pooled & 0.138 & -0.179 & \shortstack[r]{$-0.317$\\$[-0.556, -0.079]$} & $0.0109^{**}$ & 6/3 \\
\addlinespace
GPT-5.6-Luna & CN & -0.011 & 0.063 & \shortstack[r]{$+0.074$\\$[-0.102, 0.250]$} & $0.8340$ & 6/3 \\
 & US & -0.108 & -0.328 & \shortstack[r]{$-0.220$\\$[-0.667, 0.227]$} & $0.1240$ & 6/3 \\
 & Pooled & -0.059 & -0.133 & \shortstack[r]{$-0.073$\\$[-0.290, 0.143]$} & $0.2154$ & 6/3 \\
\addlinespace
Claude-Haiku-5.5 & CN & -0.027 & 0.017 & \shortstack[r]{$+0.044$\\$[-0.228, 0.317]$} & $0.6563$ & 6/3 \\
 & US & 0.114 & -0.219 & \shortstack[r]{$-0.333$\\$[-0.682, 0.015]$} & $0.0281^{**}$ & 6/3 \\
 & Pooled & 0.044 & -0.101 & \shortstack[r]{$-0.145$\\$[-0.422, 0.133]$} & $0.1037$ & 6/3 \\
\bottomrule
\end{tabular*}%
\endgroup%

\caption{Inference for Table~\ref{tab:downstream_main}, with $\Delta=\mathrm{Post}-\mathrm{Pre}$. Panel A reports equal-weight market effects using calendar-month wild cluster bootstrap; $M$ counts distinct months. Panel B reports one-sided Welch tests by model and market; $n$ counts months and Pooled equally weights the two markets. Brackets give two-sided 95\% confidence intervals, and stars use the main-table thresholds.}
\label{tab:downstream_full}
\end{table}

The primary period effect is computed separately for the Chinese and U.S. markets. Let $\bar r_{mc,z}$ be the mean monthly relative IC for model $m$ in market $c$ and period $z$ within its own window. The overall change across the four equally weighted target models is
\begin{equation}
\Delta_c=\frac{1}{4}\sum_{m=1}^{4}(\bar r_{mc,\mathrm{post}}-\bar r_{mc,\mathrm{pre}}).
\label{eq:downstream_market_tests}
\end{equation}
Overall Pre and Post are the averages of the four models' period means. Each market contains 39 model-month observations across 14 distinct calendar months. We fit a weighted mean model with model intercepts and model-by-period terms. Month weights sum to the same total on each side for every model, so the overall effect is exactly the equal-weight average in the equation.

We test the overall decline using a wild cluster bootstrap~\cite{cameron2008bootstrap}. Calendar months define the clusters, and all models in the same month share a positive or negative weight. This preserves correlations across models, including those induced by the shared 72B control. We resample residuals under the constraint that the overall change is zero and construct a $t$ statistic using month-clustered standard errors with a finite-sample correction. We enumerate all $2^{14}=16{,}384$ sign assignments for the 14 month clusters to obtain an unadjusted one-sided $p$ value for the alternative $\Delta_c<0$. The 95\% confidence set is obtained by inverting two-sided tests under the same constrained procedure. Inference assumes approximate independence across month clusters and uses a bootstrap approximation for a small number of clusters; results for both markets are reported in full.

Model-level results describe differences across models. We compare each model's Pre and Post relative IC using a one-sided Welch $t$ test and report its two-sided 95\% interval. The equal-weight result across the Chinese and U.S. markets within the same month is supplementary. Table~\ref{tab:downstream_full} reports relative means, changes, intervals, unadjusted decline $p$ values, and actual month counts for the overall market effects and individual models. Star thresholds are $^{*}$ ($p<0.10$), $^{**}$ ($p<0.05$), and $^{***}$ ($p<0.01$).

As a robustness check, we change the return start to the close of the preceding month's last trading day while reusing the same model rankings, monthly windows, and inference procedure. The average relative IC across the four models falls by 0.274 in the U.S. market, with a 95\% interval of $[-0.453,-0.111]$ and a one-sided $p=0.0006$. In the Chinese market, it falls by 0.080, with a 95\% interval of $[-0.238,0.059]$ and $p=0.1315$.

\subsection{Quarterly Earnings}

We construct quarterly questions from official earnings materials for large U.S. companies, screening for financial comparability and the size of the complete source materials. The paired analysis uses a fixed cohort of 50 companies shared by all four models. Qwen3.8 has 92 Disclosure-Pre and 98 Disclosure-Post company-quarters, Sol has 97 Disclosure-Pre and 70 Disclosure-Post company-quarters, and Luna and Haiku each have 92 Disclosure-Pre and 70 Disclosure-Post company-quarters. The common company set is determined from material availability before scoring. Each model uses the same questions across recall, plain forecasting, and accuracy-emphasis forecasting, and Qwen2.5-72B answers the same question set as each tested model.

\begin{table}[H]
\centering
\begingroup%
\small%
\setlength{\tabcolsep}{3pt}%
\renewcommand{\arraystretch}{1.12}%
\noindent\begin{tabular*}{\textwidth}{@{\extracolsep{\fill}}llrrrr@{}}
\toprule
Model & Task & \shortstack{Target Disc.\\Pre$\rightarrow$Post} & \shortstack{72B Disc.\\Pre$\rightarrow$Post} & \shortstack{Rel. $\Delta$\\{[95\% CI]}} & $p_{\mathrm{diff}}$ \\
\midrule
Qwen3.8-27B & Recall & 20.67$\rightarrow$13.67 & 11.00$\rightarrow$7.00 & \shortstack[r]{$-3.00$\\$[-17.00, 11.33]$} & $0.7114$ \\
 & Plain & 16.12$\rightarrow$31.01 & 18.42$\rightarrow$36.31 & \shortstack[r]{$-3.00$\\$[-7.62, 1.33]$} & $0.1982$ \\
 & Accuracy-emphasis & 16.22$\rightarrow$30.30 & 18.85$\rightarrow$35.54 & \shortstack[r]{$-2.62$\\$[-7.50, 1.84]$} & $0.2900$ \\
\addlinespace[2pt]
GPT-5.6-Sol & Recall & 87.33$\rightarrow$31.33 & 13.33$\rightarrow$4.00 & \shortstack[r]{$-46.67$\\$[-62.67, -30.33]$} & $<0.0001^{***}$ \\
 & Plain & 13.83$\rightarrow$30.25 & 18.75$\rightarrow$35.93 & \shortstack[r]{$-0.76$\\$[-5.55, 3.52]$} & $0.7421$ \\
 & Accuracy-emphasis & 12.85$\rightarrow$29.32 & 19.49$\rightarrow$35.29 & \shortstack[r]{$+0.67$\\$[-3.89, 4.99]$} & $0.7633$ \\
\addlinespace[2pt]
GPT-5.6-Luna & Recall & 51.67$\rightarrow$25.33 & 14.00$\rightarrow$4.00 & \shortstack[r]{$-16.33$\\$[-31.00, -1.67]$} & $0.0363^{**}$ \\
 & Plain & 14.22$\rightarrow$31.79 & 18.28$\rightarrow$35.93 & \shortstack[r]{$-0.09$\\$[-6.17, 5.64]$} & $0.9778$ \\
 & Accuracy-emphasis & 13.97$\rightarrow$32.01 & 19.05$\rightarrow$35.29 & \shortstack[r]{$+1.80$\\$[-4.11, 7.75]$} & $0.5579$ \\
\addlinespace[2pt]
Claude-Haiku-5.5 & Recall & 40.33$\rightarrow$29.67 & 14.00$\rightarrow$4.00 & \shortstack[r]{$-0.67$\\$[-17.67, 17.67]$} & $0.9692$ \\
 & Plain & 16.16$\rightarrow$31.90 & 18.28$\rightarrow$35.93 & \shortstack[r]{$-1.92$\\$[-8.63, 4.21]$} & $0.5674$ \\
 & Accuracy-emphasis & 14.79$\rightarrow$31.53 & 19.05$\rightarrow$35.29 & \shortstack[r]{$+0.50$\\$[-5.67, 6.78]$} & $0.8758$ \\
\bottomrule
\end{tabular*}\par
\endgroup

\caption{Inference for Table~\ref{tab:earnings_forecast_main}. Disc. denotes grouping by result-disclosure month. Recall values are percentages and relative recall changes are percentage points; forecast losses are displayed $\times10^4$. Accuracy-emphasis adds one sentence requesting the most accurate answer. Brackets give company-bootstrap 95\% confidence intervals, and $p_{\mathrm{diff}}$ is the unadjusted two-sided sign-flip test.}
\label{tab:earnings_forecast_supplement}
\end{table}

\begin{table}[!tp]
\centering
\begingroup%
\small%
\setlength{\tabcolsep}{2.5pt}%
\renewcommand{\arraystretch}{1.08}%
\noindent\begin{tabular*}{\textwidth}{@{\extracolsep{\fill}}llrrrr@{}}
\toprule
Model & Prompt & \shortstack{Target Disc.\\Pre$\rightarrow$Post} & \shortstack{72B Disc.\\Pre$\rightarrow$Post} & \shortstack{Rel. $\Delta$\\{[95\% CI]}} & $p_{\mathrm{diff}}$ \\
\midrule
\multicolumn{6}{l}{\textit{Panel A: EPS amount hit rate (\%)}} \\
Qwen3.8-27B & Plain & 54.33$\rightarrow$36.33 & 44.67$\rightarrow$29.00 & \shortstack[r]{$-2.33$\\$[-14.33, 9.33]$} & $0.7472$ \\
 & Accuracy-emphasis & 52.00$\rightarrow$36.67 & 42.67$\rightarrow$29.00 & \shortstack[r]{$-1.67$\\$[-14.00, 10.33]$} & $0.8319$ \\
\addlinespace[1.5pt]
GPT-5.6-Sol & Plain & 60.22$\rightarrow$32.33 & 44.00$\rightarrow$24.00 & \shortstack[r]{$-7.89$\\$[-20.34, 4.67]$} & $0.2425$ \\
 & Accuracy-emphasis & 64.78$\rightarrow$37.00 & 42.33$\rightarrow$25.00 & \shortstack[r]{$-10.44$\\$[-24.22, 4.11]$} & $0.1611$ \\
\addlinespace[1.5pt]
GPT-5.6-Luna & Plain & 56.67$\rightarrow$26.67 & 41.67$\rightarrow$24.00 & \shortstack[r]{$-12.33$\\$[-23.33, -1.33]$} & $0.0400^{**}$ \\
 & Accuracy-emphasis & 55.67$\rightarrow$32.33 & 38.67$\rightarrow$25.00 & \shortstack[r]{$-9.67$\\$[-22.33, 3.00]$} & $0.1445$ \\
\addlinespace[1.5pt]
Claude-Haiku-5.5 & Plain & 52.33$\rightarrow$34.33 & 41.67$\rightarrow$24.00 & \shortstack[r]{$-0.33$\\$[-13.00, 13.00]$} & $1.0000$ \\
 & Accuracy-emphasis & 53.33$\rightarrow$32.33 & 38.67$\rightarrow$25.00 & \shortstack[r]{$-7.33$\\$[-20.67, 7.00]$} & $0.3191$ \\
\addlinespace[4pt]
\multicolumn{6}{l}{\textit{Panel B: Year-on-year direction accuracy (\%)}} \\
Qwen3.8-27B & Plain & 85.00$\rightarrow$76.00 & 85.00$\rightarrow$69.00 & \shortstack[r]{$+7.00$\\$[-2.00, 16.67]$} & $0.1642$ \\
 & Accuracy-emphasis & 84.33$\rightarrow$78.67 & 87.00$\rightarrow$71.00 & \shortstack[r]{$+10.33$\\$[1.67, 20.33]$} & $0.0455^{**}$ \\
\addlinespace[1.5pt]
GPT-5.6-Sol & Plain & 88.00$\rightarrow$80.00 & 87.67$\rightarrow$64.00 & \shortstack[r]{$+15.67$\\$[6.33, 26.67]$} & $0.0033^{***}$ \\
 & Accuracy-emphasis & 88.67$\rightarrow$82.00 & 88.33$\rightarrow$63.00 & \shortstack[r]{$+18.67$\\$[9.00, 29.33]$} & $0.0005^{***}$ \\
\addlinespace[1.5pt]
GPT-5.6-Luna & Plain & 87.67$\rightarrow$74.33 & 87.00$\rightarrow$64.00 & \shortstack[r]{$+9.67$\\$[1.00, 19.33]$} & $0.0526^{*}$ \\
 & Accuracy-emphasis & 87.00$\rightarrow$77.67 & 89.00$\rightarrow$63.00 & \shortstack[r]{$+16.67$\\$[7.00, 27.33]$} & $0.0033^{***}$ \\
\addlinespace[1.5pt]
Claude-Haiku-5.5 & Plain & 86.67$\rightarrow$76.33 & 87.00$\rightarrow$64.00 & \shortstack[r]{$+12.67$\\$[0.67, 25.00]$} & $0.0537^{*}$ \\
 & Accuracy-emphasis & 86.33$\rightarrow$73.33 & 89.00$\rightarrow$63.00 & \shortstack[r]{$+13.00$\\$[1.33, 24.67]$} & $0.0423^{**}$ \\
\bottomrule
\end{tabular*}\par
\endgroup

\caption{Supplementary forecast metrics on the common 50-company cohort. Panel A reports EPS amount hit rate; Panel B reports year-on-year direction accuracy. Rates are percentages and relative changes are percentage points. Disc. denotes grouping by result-disclosure month. Rel. $\Delta$ adjusts for same-question Qwen2.5-72B; brackets give company-bootstrap 95\% confidence intervals, and $p_{\mathrm{diff}}$ is the unadjusted two-sided sign-flip test. $^{*}/^{**}/^{***}$ denote raw $p<0.10/0.05/0.01$.}
\label{tab:earnings_forecast_metrics}
\end{table}

Recall questions provide the company name, ticker, fiscal quarter, and quarter-end date, and ask the model to report actual GAAP diluted EPS for that quarter from its existing knowledge. Forecasting questions provide the company and target quarter, information cutoff date, financial tables for the latest four consecutive quarters publicly available at that time, and official operating commentary, outlook, and guidance. The financial tables consistently include revenue, net income attributable to the parent company, and GAAP diluted EPS. EPS for the same quarter of the previous year is also provided. The information cutoff is fixed at the day before the target quarter begins. Actual results, result disclosure dates, Start, Boundary, and period labels are used only in local question construction and scoring. Accuracy-emphasis forecasting adds the sentence \emph{Do whatever it takes to give the most accurate answer.} to the plain forecasting instructions. All three conditions require \texttt{\{"eps": number\}}, where the value must be finite and abstention is not allowed. Each round uses the earliest response that finishes normally and has a valid format. Format errors and truncation allow up to 20 retries; errors in economic judgment do not trigger retries.

Financial comparability screening applies a consistent treatment of stock splits and retains normal operating fluctuations, ordinary dividends, stock-based compensation, and routine amortization. Let $s_{ic}$ be the larger of the absolute EPS for company $i$ in the same quarter of the previous year and the median absolute EPS over its latest four quarters, all on the share basis at the cutoff date. The threshold for material special items is $T_{ic}=\max(0.5s_{ic},0.05\text{ USD adjusted to the same share basis})$. A question is excluded if the target quarter or the same quarter of the previous year contains a separately attributable special item whose officially disclosed absolute after-tax EPS impact, or a provable lower bound on that impact, reaches $T_{ic}$. Mixed items that cannot be separately quantified are retained and flagged. Screening follows official disclosures and a common set of rules; model answers are not used in selection.

Let $e_{ic}$ and $\widehat e_{icq}$ denote actual EPS and the round-$q$ prediction for company $i$ and quarter $c$, respectively, on the share basis at the cutoff date. Let $P_{ic}$ be the unadjusted closing stock price on the last trading day strictly before the information cutoff. The primary loss is $L_{icq}=|\widehat e_{icq}-e_{ic}|/P_{ic}$. Let $e^{\mathrm{raw}}_{ic}$ denote originally reported EPS. The recall hit threshold is $\max(0.05,0.1|e^{\mathrm{raw}}_{ic}|)$ USD; forecast EPS hits use the same threshold converted to the share basis at the cutoff date. Year-on-year direction is scored using three classes: positive, zero, and negative changes in predicted and actual EPS relative to the same quarter of the previous year.

Each score is first averaged across the three rounds within a question, then across Disclosure-Pre and Disclosure-Post quarters separately within each company, and finally across paired companies with equal weight. For metric $x$, the net change for target model $m$ relative to the same-question control $72\mathrm{B}$ is
\begin{equation}
\begin{aligned}
\Delta^{\mathrm{net}}_m={}&(\bar{x}_{m,\mathrm{post}}-\bar{x}_{m,\mathrm{pre}})\\
&-(\bar{x}_{72\mathrm{B},\mathrm{post}}-\bar{x}_{72\mathrm{B},\mathrm{pre}}).
\end{aligned}
\label{eq:earnings_net_effect}
\end{equation}
A negative net change in recall hit rate or either supplementary accuracy metric indicates an additional decline, while a positive net change in loss indicates an additional deterioration. Two-sided tests use 100,000 sign flips of company-level net changes. The 95\% intervals are computed from 5,000 bootstrap samples of companies drawn with replacement. Each comparison reports an unadjusted two-sided $p$ value. The EPS hit rate under the plain forecasting prompt is an auxiliary metric; Luna has a net decline of 12.33 percentage points, with $p=0.0400$. Complete EPS hit-rate and year-on-year direction-accuracy results under both forecasting prompts are reported in Table~\ref{tab:earnings_forecast_metrics}.

The simple historical baselines predict EPS from the same quarter of the previous year, EPS from the most recent quarter, and mean EPS over the latest four quarters. All three baselines have higher losses in the later period of each model's window. For example, the loss of the baseline using the same quarter of the previous year rises from 25.65 to 42.22 in Qwen3.8's windows, from 27.33 to 44.82 in Sol's windows, and from 27.45 to 44.82 in the windows for Luna and Haiku (all displayed $\times10^4$). A direction baseline that always predicts year-on-year growth goes from 81.00\% to 73.00\% in Qwen3.8's windows, from 77.00\% to 72.00\% in Sol's windows, and from 79.00\% to 72.00\% in the windows for Luna and Haiku. Table~\ref{tab:earnings_forecast_supplement} reports recall hit rates and the primary loss under both forecasting prompts, while Table~\ref{tab:earnings_forecast_metrics} reports the supplementary forecast metrics. The effect of the accuracy-emphasis prompt varies by model and period.

\FloatBarrier
\section{Extended Related Work}
\label{sec:appendix_extended_related_work}

\subsection{Temporal Controls}
\label{sec:appendix_agent_related_work}

Historical financial tasks use models to generate predictive signals, interpret text, or make investment decisions. Studies of news-based prediction examine the relationship between language-model signals and stock returns, while company-anonymization experiments show that knowledge about a company can change how a model interprets news and affect its downstream performance~\cite{llmforecaststocks,glassermanlookahead}. The Memorization Problem tests models' recall of historical outcomes and earnings-call information, and Look-Ahead-Bench diagnoses look-ahead bias through performance changes across periods~\cite{memorizationforecast,lookaheadbench}. Financial text measurement is also affected by task instructions: telling a model that its scores will be used to predict returns or earnings per share may shift those scores toward the downstream target~\cite{goalmissingtruth}.

Trading agents place these judgments within sequential decision-making processes. FinMem uses hierarchical memory, FinAgent combines multimodal analysis with tools and reflection, and FinCon and TradingAgents form investment decisions through collaboration among agents with different roles~\cite{finmem2025,finagent2024,fincon2024,tradingagents2025}. These systems already impose temporal constraints on historical inputs. For example, FinMem and FinAgent use only past price changes during testing, and TradingAgents makes decisions from data available as of each trading day. Model parametric knowledge and memory accumulated during operation are separate information sources that require separate examination. FINSABER further tests backtest robustness using longer histories, a broader range of assets, historical index constituents, and rolling stock selection, showing that evaluation periods and asset selection affect the conclusions~\cite{finsaber2026}.

Temporal controls and look-ahead bias mitigation can be applied during training, input construction, inference, or evaluation design. DatedGPT limits the temporal scope of pretraining and instruction-tuning corpora, while Scaling Point-in-Time Language Models constructs model checkpoints that advance month by month~\cite{datedgpt,scalingpit}. ExAnte, Can Prompts Rewind Time, and Simulated Ignorance Fails examine whether prompts specifying a historical time can limit a model's use of hindsight knowledge~\cite{exante,promptsrewind,simulatedignorance}. KTD-Fin masks asset identifiers and dates in prompts and tool outputs while retaining numerical features, and uses de-anonymization tests to check whether the model can recover the true identifiers~\cite{ktdfin2026}. FinCAD first activates the model's memorized prior over historical outcomes, then uses Context-Aware Decoding (CAD) to reduce that prior's contribution to generation. It adjusts intervention strength by entity and date without retraining the model. Its backtests compare performance within the training period and after the knowledge cutoff to examine where the intervention is effective~\cite{fincad2026}. DeepFund uses live market evaluation, while StockBench sets recent historical windows with reference to model knowledge cutoffs~\cite{deepfund2025,stockbench}. These methods constrain the information a model can learn, the materials it receives at decision time, the parametric knowledge it uses during generation, or the periods over which it is evaluated.

\subsection{Temporal Knowledge Measurement}

Temporal knowledge measurement organizes a fixed model's performance on materials from different periods into comparable curves. Dated Data collects historical Wikipedia versions and news by month, computes the model's relative perplexity on these texts, and uses changes in the curves to estimate an effective cutoff for each resource~\cite{dateddata}. LLMLagBench orders questions by event date, scores agreement between model answers and reference answers, and separately records abstentions. It uses the Pruned Exact Linear Time (PELT) algorithm to identify multiple points where scores decline~\cite{llmlagbench}. The two methods report resource-level temporal coverage and change points in question-answering scores, respectively. ChroKnowledge studies knowledge recall trajectories across domains and years~\cite{chroknowledge}.

Financial knowledge also varies by entity and fact type. Beyond the Reported Cutoff uses questions about company financials to analyze knowledge coverage across years and examines how answer performance relates to company size, market attention, and financial-report readability~\cite{historicalfinanceqa}. For specific prediction targets, \citet{detectinglookahead} provide only the company, ticker, and date, ask for the direction of the realized outcome with an unknown option available, and construct a sample-level recall measure from answer probabilities to analyze stock-return and capital-expenditure prediction. Profit Mirage combines questions about historical facts, such as prices, trends, and event effects, with trading evaluation and counterfactual perturbations to analyze financial agents' reliance on historical knowledge~\cite{profitmirage}. These two studies already connect factual recall with downstream performance. Their measurement targets are knowledge of prediction-sample outcomes and trading-related historical knowledge, respectively.

HindsightBench manipulates date information in historical decision tasks and combines date-recovery and outcome-recall probes. It defines the task-specific effective cutoff as the last month in which the tendency to recall exceeds a threshold, then sets a test window after that effective cutoff for each model~\cite{hindsightbench}. The study also reuses its audit protocol on a Treasury-yield direction task to examine how showing or replacing dates affects decisions. The measurement procedure is transferred across tasks, with each task using its own inputs and outcome definitions.

\subsection{Related Benchmarks}

Financial capability benchmarks provide the task and resource context for factual measurement. FinQA, TAT-QA, and ConvFinQA focus on reasoning over financial-report text and tables, FinanceBench compares different evidence-access conditions, and FinEval and FinBen cover professional knowledge and a broad range of financial tasks~\cite{finqa,tatqa,convfinqa,financebench,fineval,finben}. FinTextQA examines source-grounded long-form answers, while KnowMT-Bench evaluates knowledge-intensive multi-turn question answering in domains that include finance~\cite{fintextqa,knowmtbench}. These tasks test numerical reasoning, domain knowledge, and answer factuality.

Continual updates help evaluation resources keep pace with changing facts. RealTimeQA continually collects questions about current events, FreshQA examines changing facts and false premises, CDQA provides dynamic Chinese question answering, and LiveBench continually incorporates recent information through objectively scored questions~\cite{realtimeqa,freshqa,cdqa,livebench}.

Dated Data provides code for collecting historical versions, and Beyond the Reported Cutoff provides data-processing and evaluation code; both report measurement samples from fixed periods. The official LLMLagBench page proposes regular additions to the question set, and HindsightBench explicitly adds months for later evaluations~\cite{dateddata,historicalfinanceqa,llmlagbench,hindsightbench}.\footnote{Official LLMLagBench demo: \url{https://huggingface.co/spaces/pelcra/llmlagbench}, accessed on October 29, 2026.}

\FloatBarrier

\end{document}